%% file: arxiv.tex
\documentclass[acmtog,nonacm,screen]{acmart}
\acmSubmissionID{}

\input{preamble}
\arxiv

\usepackage{booktabs} 

\defcitealias{hunyuan3domni}{Tencent 2025b}

\usepackage[ruled]{algorithm2e} 

\SetAlFnt{\small}
\SetAlCapFnt{\small}
\SetAlCapNameFnt{\small}
\SetAlCapHSkip{0pt}

\begin{document}
\title{Make-It-Poseable: Feed-forward Latent Posing Model for 3D Characters}

\author{Zhiyang Guo}
\orcid{0000-0002-4278-1349}
\email{guozhiyang@mail.ustc.edu.cn}
\affiliation{%
  \institution{University of Science and Technology of China}
  \city{Hefei}
  \country{China}
}

\author{Ori Zhang}
\email{cuminflea@gmail.com}
\affiliation{%
  \institution{Independent Researcher}
  \country{China}
}

\author{Jax Xiang}
\affiliation{%
  \institution{Independent Researcher}
  \country{China}
}

\author{Alan Zhao}
\affiliation{%
  \institution{Independent Researcher}
  \country{China}
}

\author{Zhenxun Yuan}
\orcid{0009-0000-2674-7434}
\email{yuanzhenxun@mail.ustc.edu.cn}
\affiliation{%
  \institution{University of Science and Technology of China}
  \city{Hefei}
  \country{China}
}

\author{Wengang Zhou}
\authornote{Corresponding author.}
\orcid{0000-0003-1690-9836}
\email{zhwg@ustc.edu.cn}
\affiliation{%
  \institution{University of Science and Technology of China}
  \city{Hefei}
  \country{China}
}

\author{Houqiang Li}
\orcid{0000-0003-2188-3028}
\email{lihq@ustc.edu.cn}
\affiliation{%
  \institution{University of Science and Technology of China}
  \city{Hefei}
  \country{China}
}

\renewcommand\shortauthors{Guo, Z. et al}

\begin{teaserfigure}
    \centering
    \includegraphics[width=1.0\linewidth, trim = 0cm 0cm 0cm 0cm, clip]{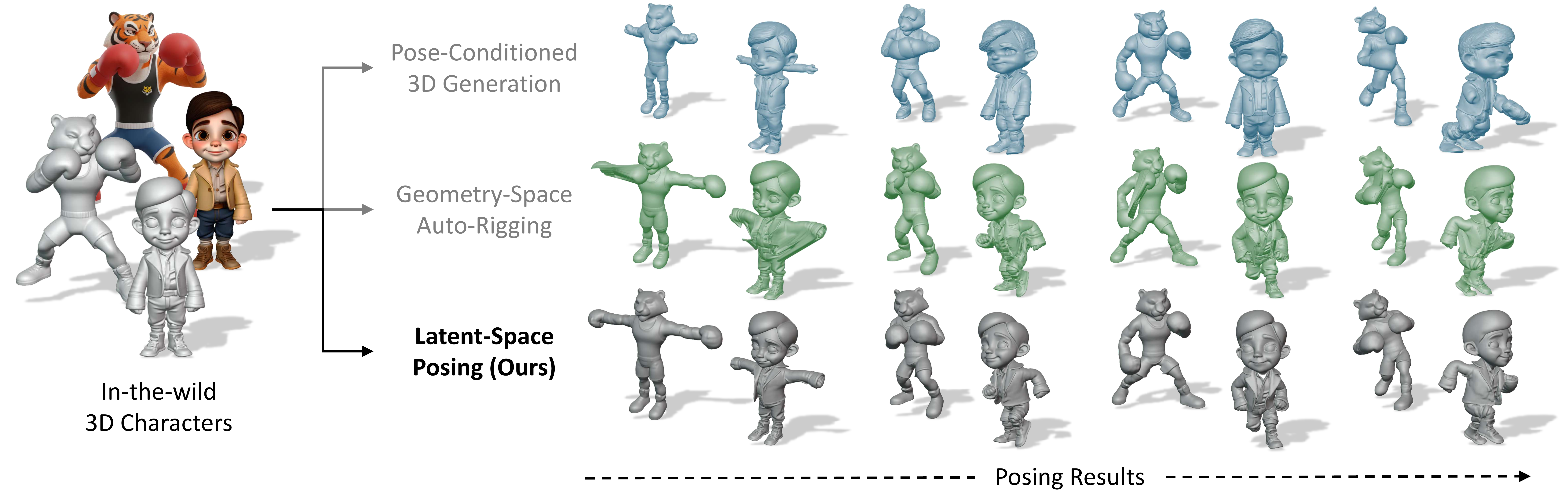}
    \captionof{figure}{Given a 3D character model of arbitrary shape and initial pose, our method efficiently re-poses it in a single feed-forward pass. Unlike auto-rigging or generative approaches that often suffer from skinning artifacts or limited controllability, our latent posing paradigm robustly handles challenging cases and produces high-fidelity results.}
    \Description{Teaser}
    \label{fig:teaser}
\end{teaserfigure}

\begin{abstract}
Posing 3D characters is a fundamental task in computer graphics.
However, existing paradigms, ranging from traditional auto-rigging to recent pose-conditioned generative models, frequently struggle with inaccurate skinning weights, fixed mesh topologies, and poor pose conformance. These challenges have become particularly pronounced with the recent explosion of AI-generated 3D assets, which often exhibit flawed structures and fused geometry.
To address these issues, we introduce \textbf{Make-It-Poseable}, a novel feed-forward framework that reformulates character posing as a skinning-free latent-space transformation problem.
By decoupling shape deformation from the constraints of fixed mesh connectivity, our method directly operates on compact latent representations to reconstruct characters in target poses.
To achieve this, our framework integrates a latent posing transformer for shape manipulation, a dense pose representation for fine-grained control, and an adaptive completion module optimized via a bipartite-matched latent loss to robustly handle topological changes.
Extensive experiments demonstrate that our method significantly outperforms existing baselines in posing quality. 
Furthermore, our design shows promising generalization to diverse morphologies such as quadrupeds in our qualitative tests, and supports various 3D authoring applications such as part replacement and refinement.
\end{abstract}


%
%


\maketitle

\input{samplebody-journals}

\bibliographystyle{ACM-Reference-Format}
\bibliography{sample-bibliography}

\clearpage

\input{supplement}

\end{document}

%% file: preamble.tex
\usepackage{multirow}
\usepackage{bm}

\allowdisplaybreaks[4]

\newcommand{\parsection}[1]{\noindent\textbf{#1.}~}

\newif\ifreview 
\newif\ifarxiv \newcommand{\arxiv}{\arxivtrue}
\newif\ifcamera 
\newif\ifrebuttal 

\RequirePackage{xspace}
\makeatletter
\DeclareRobustCommand\onedot{\futurelet\@let@token\@onedot}
\def\@onedot{\ifx\@let@token.\else.\null\fi\xspace}

\def\eg{\emph{e.g}\onedot} 

\def\ie{\emph{i.e}\onedot}

\makeatother

%% file: samplebody-journals.tex
\section{Introduction}
\label{sec:intro}

Posing 3D characters, \ie, skeleton-guided shape deformation aimed at controllably articulating those static meshes into novel poses, is a fundamental task in computer graphics.
Beyond its role in final animation, robust character posing is increasingly essential for 3D asset authoring and animation preprocessing.
Recently, as AI-generated content emerges as a pivotal source of 3D assets~\cite{zhang2024clay,li2025triposg,hunyuan3d21}, the demand for high-quality and efficient posing of them has surged.

However, traditional posing pipelines based on rigging and skinning face significant hurdles when faced with these ``in-the-wild'' assets.
Manual rigging is notoriously labor-intensive, and even state-of-the-art auto-rigging methods~\cite{MIA,liu2025riganything,song2025magicarticulate,zhang2025unirig,song2025puppeteer} frequently fail to reach production-level skinning quality, suffering from artifacts due to weight leakage.
More critically, AI-generated meshes often exhibit topological imperfections, such as fused limbs or self-occlusions. Since traditional geometry-based approaches are constrained by a fixed mesh topology, they are fundamentally incapable of resolving these defects or handling deformations that require topological changes like revealing newly exposed surfaces.

Alternatively, space-warping methods like Neural Generalized Cylinder~\cite{NGC} attempt to bypass fixed mesh connectivity by deforming implicit neural fields.
Nevertheless, this approach relies heavily on manual cylinder configuration and struggles to maintain high-fidelity surface details or synthesize plausible geometry for newly exposed regions.
Another line of work explores pose-conditioned 3D generation~\cite{yan2025posemaster,hunyuan3domni}.
While capable of accommodating topological shifts, these generative methods often involve slow iterative denoising processes and sparse pose signals, leading to insufficient pose conformance and inherent difficulty in preserving the identity of the source character.

In this paper, we introduce \textbf{Make-It-Poseable}, a novel feed-forward framework that reformulates character posing as a latent-space transformation problem.
Instead of deforming vertices or warping space, we directly manipulate the compact latent representation of 3D shapes to reconstruct the character in the target pose.
Our approach is inherently skinning-free and topology-aware, allowing it to robustly re-pose characters with complex self-contacts or fused geometry.
By sidestepping the aforementioned limitations of geometry-space manipulation and cross-modal generation, our method can serve as an effective geometry authoring and refinement tool for modern 3D production pipelines.
To the best of our knowledge, this is the \emph{first} feed-forward 3D character posing model that operates on a native 3D latent representation.

The core of our framework is a \emph{latent posing transformer} that predicts the re-posed character's latent VecSet~\cite{zhang20233dshape2vecset} shape representation in a single feed-forward pass.
Embracing minimal skinning inductive bias, our model bypasses explicit, error-prone skinning weights, learning instead to manipulate shapes directly within the latent space in response to skeletal articulation.
To facilitate this complex manipulation, we address the spatial alignment ambiguity of unordered VecSets by introducing a \emph{dense pose representation}.
Unlike sparse pose conditioning that inherently creates information bottlenecks~\cite{yan2025posemaster,hunyuan3domni}, our model injects spatially aligned, fine-grained skeletal context into the shape latents.
This explicit dense correspondence not only enables precise deformation control, but also establishes a canonical transformation path, unlocking a direct \emph{latent-space supervision} strategy that prevents the high-frequency detail loss.

Furthermore, when posing assets with initial topological flaws (\eg, fused limbs), large deformations inevitably expose missing geometry, necessitating the on-the-fly generation of additional shape latents.
To address this, we introduce an \emph{adaptive completion module} during a dedicated fine-tuning stage.
Rather than resorting to geometry-space supervision that incurs heavy computational overhead and spatial ambiguity, we maintain a latent-space approach via a novel anchor-guided mechanism.
Specifically, the module first explicitly predicts 3D spatial anchors for newly exposed regions, which then guide the generation of supplementary latents.
By aligning these predicted and ground-truth anchors via the Hungarian algorithm, we optimize this module efficiently through a bipartite-matched latent loss.
This design effectively guarantees plausible posing results of our framework.

Extensive experiments demonstrate that our method significantly outperforms existing approaches in posing quality, while naturally supporting other 3D shape editing applications, such as the rescaling, segmentation, replacement, and refinement of body parts.
Crucially, while still relying on an extracted skeletal prior, our framework is intrinsically agnostic to the skeleton template rather than tied to a fixed humanoid rig.
Despite being trained exclusively on bipedal data---a choice made to leverage large-scale animation data and to tackle severe self-contact scenarios unique to humanoids---our model exhibits promising zero-shot generalization to out-of-distribution inputs on the tested examples.
It articulates diverse, unseen morphologies, including characters with novel accessories and quadrupedal animals, without any category-specific fine-tuning.

\section{Related Works}
\label{sec:related}

\subsection{Latent 3D Generation and Editing}

Modern 3D generation increasingly operates in latent space to achieve superior geometric generalization.
While early approaches reconstruct geometry by lifting 2D multi-view features~\cite{poole2022dreamfusion, liu2023zero, long2024wonder3d}, recent works shift toward native 3D architectures, representing shapes as unordered VecSets~\cite{zhang20233dshape2vecset, li2025triposg, hunyuan3d21} or sparse voxel grids~\cite{xiang2025trellis, ren2024xcube}.
These representations facilitate diverse editing operations, including part-aware composition~\cite{yang2025holopart, yan2025x, zhang2025bang}, geometry enhancement~\cite{deng2025detailgen3d, ye2025hi3dgen}, and localized refinement~\cite{li2025voxhammer, li2025relate3d}.
Beyond direct editing, other works interactively explore the latent space, \eg, via coupled vision-language and shape spaces~\cite{hu2023clipxplore} or human-in-the-loop subspace search~\cite{chiu2020human}.
Notably, operating as an intermediate paradigm between explicit geometry-space and pure latent-space manipulations, Neural Generalized Cylinder~\cite{NGC} introduces a space-warping approach for skeleton-like control. However, it remains incapable of separating fused artifacts or synthesizing newly exposed structures---challenges our latent posing framework is explicitly designed to overcome.

\begin{figure*}[t]
    \centering
    \includegraphics[width=0.95\linewidth]{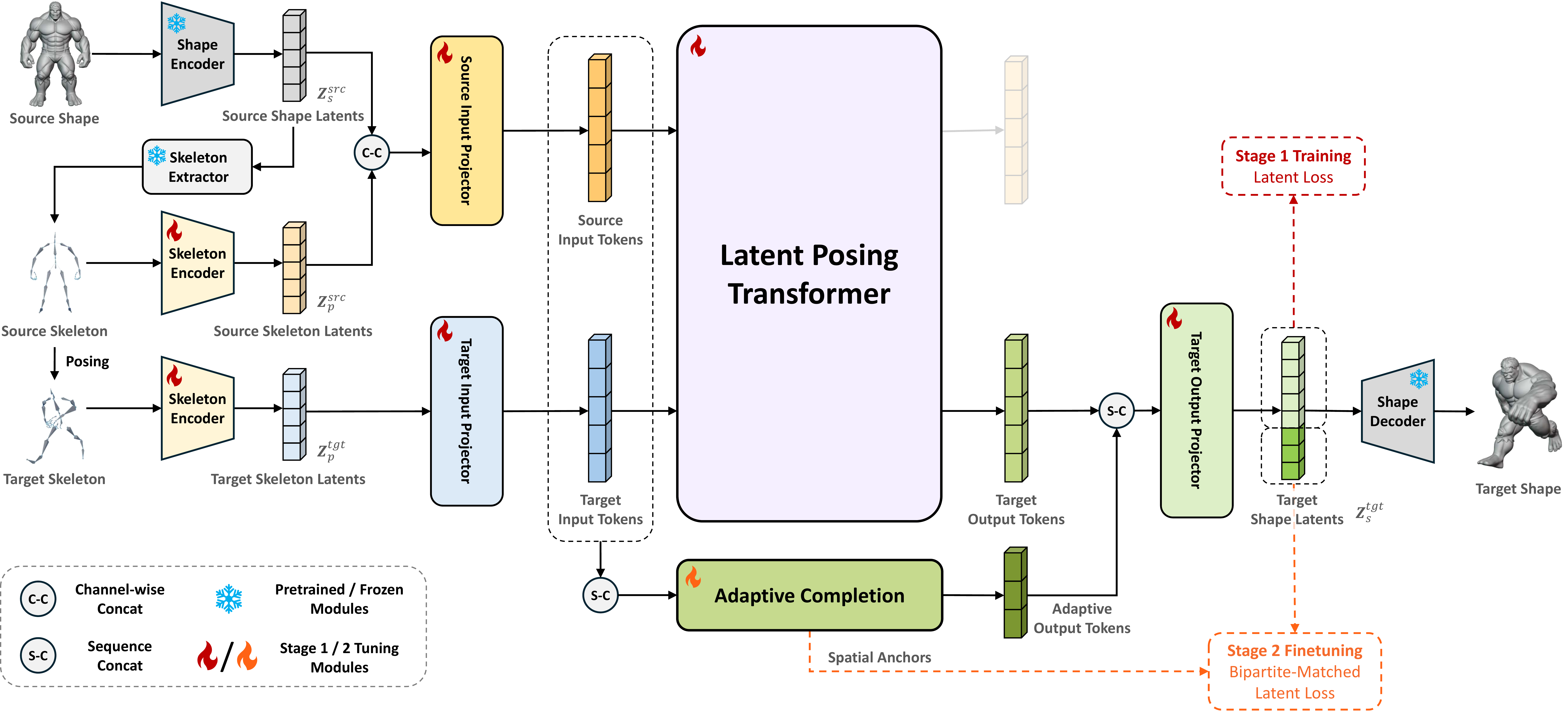}
    \caption{\textbf{Pipeline of our latent posing framework.}
    \Description{Pipeline of our latent posing framework.}
    Given a source character, we encode it into shape latents, from which a pretrained skeleton extractor predicts the source skeleton.
    Both the source and posed target skeletons are then mapped into corresponding latent representations by a shared encoder.
    Guided by this shape and skeletal context, a latent posing transformer predicts target shape latents, which are subsequently decoded into the posed mesh.
    Training proceeds in two stages.
    First, a latent loss is applied to preserve fine geometric details. Second, an anchor-guided adaptive completion module is fine-tuned via a bipartite-matched latent loss to synthesize newly exposed structures arising from necessary topological changes during deformation.
    }
    \label{fig:pipeline}
\end{figure*}

\subsection{Character Posing}

Posing 3D characters aims to articulate static meshes into novel, controllable configurations. The most prevalent approach in traditional pipelines relies on rigging and skinning, where an input mesh is deformed via a skeleton and associated blend weights.

The notoriously labor-intensive nature of manual rigging has driven the development of automatic techniques.
Traditional pipelines rely on embedding skeletons and optimizing skinning weights~\cite{baran2007automatic, xu2020rignet, ma2023tarig}.
More recently, feed-forward transformer architectures have enabled fast and robust auto-rigging for humanoids and general articulated objects~\cite{MIA, chu2024humanrig, sun2025drive, song2025magicarticulate, liu2025riganything, deng2025anymate, zhang2025unirig, song2025puppeteer}.

Despite these advances, existing data-driven auto-rigging algorithms exhibit notable limitations.
Relying on a fixed mesh topology, they inherently struggle with non-rest complex input poses or AI-generated assets that frequently contain topological imperfections (\eg, fused limbs or self-occlusions).
Furthermore, they often suffer from skinning weight leakage across geometrically proximal vertices, leading to severe deformation artifacts.

Alternatively, character posing can be formulated through the lens of pose-conditioned 3D generation.
Recent approaches~\cite{yan2025posemaster,hunyuan3domni} manage to integrate skeletal conditioning into 3D generative pipelines.
However, they often struggle with precise and generalizable pose control due to the sparse and complex mapping between skeletal signals and dense 3D geometry.
In parallel, template-based frameworks~\cite{liao2024tada, huang2024dreamwaltzg} employ canonical human templates~\cite{loper2015smpl,pavlakos2019smplx} alongside learned deformation fields to maintain structural consistency. However, this heavy reliance on human-specific templates inherently restricts their generalizability to highly stylized characters.
More recent work~\cite{wang2025wonderhuman} unifies geometry, pose, and appearance generation within a feed-forward paradigm.
Related efforts also address cross-topology motion transfer~\cite{chen2025motion2motion}.
Nevertheless, current pose-conditioned 3D generation models generally fail to achieve an ideal balance between high-fidelity detail preservation and strict pose conformance.

\subsection{Feed-forward Paradigms in 3D}

Recent feed-forward models have enabled single-pass reconstruction of high-quality 3D assets~\cite{hong2023lrm, tang2024lgm, zhang2024gslrm}, and have also been extended to dense geometry estimation and novel view synthesis~\cite{wang2024dust3r, wang2025vggt, jin2025lvsm}.
Despite this rapid progress in static 3D domains (see survey~\cite{zhang2025advances}), their application in pose-oriented dynamic 3D content remains underexplored.
While HumanRAM~\cite{yu2025humanram} represents one of the first attempts to endow LRM with animation capabilities in the 2D image domain, efficiently integrating pose understanding and geometric deformation control into native 3D feed-forward architectures remains an open challenge.

\section{Method}
\label{sec:method}

Our goal is to re-pose a given 3D character mesh while maintaining fine geometric details and shape plausibility.
To achieve this, we reformulate character posing as a latent-space transformation problem.
An overview of our framework is illustrated in Fig.~\ref{fig:pipeline}.
We first detail the shape latent space and the extraction of skeleton (Sec.~\ref{sec:preliminaries}).
Next, we introduce the shared skeleton encoder (Sec.~\ref{sec:skeleton}) and the latent posing transformer (Sec.~\ref{sec:transformer}), which jointly predict the target shape.
Finally, we describe our two-stage training paradigm: the latent-space supervision (Sec.~\ref{sec:training}), followed by the fine-tuning of an adaptive completion module to synthesize newly exposed geometry (Sec.~\ref{sec:adaptive_tokens}).
More details are provided in the supplementary material.

\subsection{Preliminaries}
\label{sec:preliminaries}

\parsection{3D VAE}
We employ a 3D Variational Autoencoder (VAE) to obtain compact, VecSet-based representation~\cite{zhang20233dshape2vecset} of 3D shapes.
Specifically, the \textit{shape encoder} takes a surface point cloud $\bm{P} \in \mathbb{R}^{N \times 3}$ as input and uses Farthest Point Sampling (FPS) to select a subset of $L$ query anchors $\tilde{\bm{P}} \in \mathbb{R}^{L \times 3}$.
Through cross-attention, these queries interact with the full set $\bm{P}$ to predict a Gaussian distribution parameterized by mean $\operatorname{E}(\bm{Z}_s)$ and variance $\operatorname{Var}(\bm{Z}_s) \in \mathbb{R}^{L \times d}$, from which the latent VecSet $\bm{Z}_s \in \mathbb{R}^{L \times d}$ is sampled:
\begin{equation}
    \operatorname{E}(\bm{Z}_s), \operatorname{Var}(\bm{Z}_s) = \operatorname{CrossAttn}(\tilde{\bm{P}}, \bm{P}),
\label{eq:encoder}
\end{equation}
where the concatenated normal input and Fourier positional encoding are omitted for simplicity.
Even without explicit constraints, each latent vector in $\bm{Z}_s$ naturally captures the local geometric context around its corresponding anchor~\cite{deng2025detailgen3d}, bridging geometry and latent spaces.
Finally, a transformer-based \textit{shape decoder} predicts the Signed Distance Function (SDF) conditioned on $\bm{Z}_s$, enabling explicit mesh reconstruction via marching cubes.

\parsection{Skeletal prior}
Traditionally, character posing relies on a skeleton $\bm{S} \in \mathbb{R}^{K \times 6}$ and blend weights for explicit vertex deformation.
We instead propose an end-to-end skinning-free latent-space solution where the skeleton $\bm{S}$ serves purely as an intuitive pose proxy to guide latent transformations.
Specifically, we utilize MIA~\cite{MIA} as our skeleton extractor, as it natively operates on VecSet representations.
To ensure seamless integration, we re-train MIA to align its latent space with our higher-capacity shape VAE.
This alignment not only allows us to directly use shape latents as inputs, bypassing redundant 3D encoding steps, but also leverages our VAE's larger capacity to improve the fidelity of the skeletal prior.

\subsection{Skeleton Encoder}
\label{sec:skeleton}

\begin{figure*}[t]
    \centering
    \includegraphics[width=0.9\linewidth]{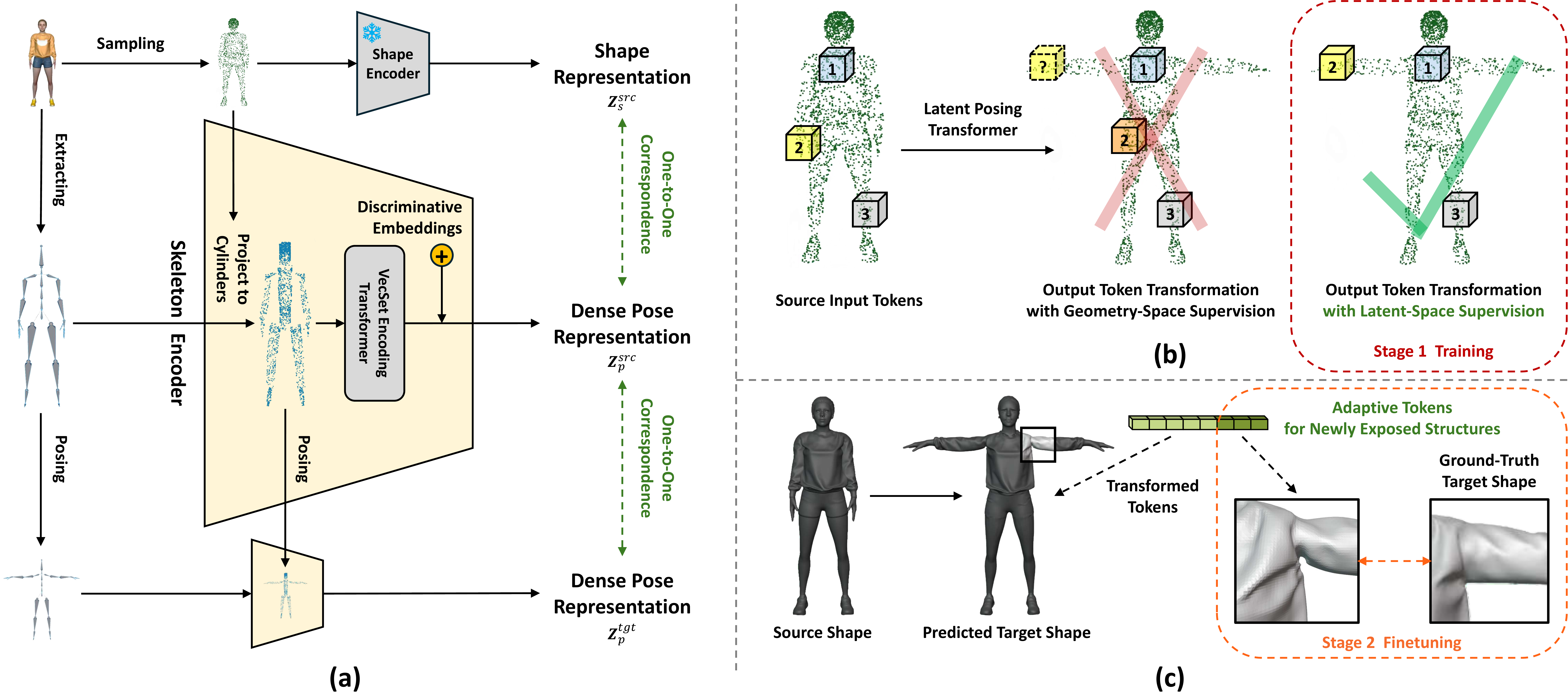}
    \caption{\textbf{Illustration of our key designs.} (a) The skeleton encoder (Sec.~\ref{sec:skeleton}) produces dense pose representations with latent-level one-to-one correspondence. (b) Latent-space supervision (Sec.~\ref{sec:training}) ensures a semantically meaningful token transformation path to preserve geometric details. (c) Adaptive completion (Sec.~\ref{sec:adaptive_tokens}) is introduced in the fine-tuning stage to handle newly exposed structures after posing.}
    \Description{Illustration of our key designs.}
    \label{fig:tokens}
\end{figure*}

As illustrated in Fig.~\ref{fig:tokens}~(a), to effectively condition the 3D shape on pose transformation, we propose a skeleton encoder that densifies the sparse skeletal representation to match the per-point granularity of the input.

\parsection{Dense pose representation}
Original sparse skeletal signals create information bottlenecks and impose a significant learning burden, as transformers must implicitly learn the complex mappings from a global pose to its spatially varying influence on every local 3D point.
Inspired by multi-view vision encoders where per-pixel Pl\"ucker ray embeddings are preferred over sparse camera matrices, we alleviate this issue by providing per-point dense skeletal context.

Specifically, each sampled point in $\bm{P}$ (\ie, the input of the shape encoder) is geometrically projected onto the surface of its associated bone. This bone, modeled as a cylinder with a radius proportional to its length, is identified via the maximum influence from the coarse blend weights provided by our pretrained skeleton extractor (though simpler segmentation priors could also be used).

These projected coordinates are then processed by a skeleton encoder that mirrors the shape encoder's architecture, except for its deterministic output.
By sharing the subsampling indices from the shape encoding process, the skeleton encoder ensures that its output latents $\bm{Z}^{src}_p \in \mathbb{R}^{L \times d}$ maintain a strict one-to-one correspondence with the shape latents $\bm{Z}^{src}_s$. This enables the effective channel-wise concatenation of fine-grained dense pose context.
Overall, the source skeleton encoding process is formulated as:
\begin{equation}
    \bm{Z}^{src}_p = \operatorname{CrossAttn}(\operatorname{Proj}(\tilde{\bm{P}}), \operatorname{Proj}(\bm{P})),
\label{eq:source_skeleton}
\end{equation}
where $\operatorname{Proj}(\cdot)$ denotes the geometric projection onto the associated bones of the source skeleton.

To establish robust source-target correspondence, the target skeleton is encoded into $\bm{Z}^{tgt}_p$ via the shared skeleton encoder. 
Crucially, instead of independently re-sampling or re-projecting target points---which would destroy latent alignment---we directly apply the bone-wise rigid deformation $\mathcal{T}(\cdot)$ to both the source fullset and subset projected points:
\begin{equation}
    \bm{Z}^{tgt}_p = \operatorname{CrossAttn}(\mathcal{T}(\operatorname{Proj}(\tilde{\bm{P}})), \mathcal{T}(\operatorname{Proj}(\bm{P}))),
\label{eq:target_skeleton}
\end{equation}
By explicitly transforming the source coordinates, this approach maintains latent-level correspondence and ensures the interpretability of target latents for subsequent supervision (Sec.~\ref{sec:training}).

\parsection{Discriminative embeddings}
While our skeleton encoding (Eqs.~\ref{eq:source_skeleton} and~\ref{eq:target_skeleton}) inherently preserves one-to-one source-target correspondence, we find in practice that this implicit 3D context alone is insufficient for the transformer to reliably preserve fine geometric details during articulation.
To explicitly highlight this correspondence and provide a ``shortcut'' for the network to transform source geometries to target locations, we introduce a set of unique learnable embeddings to build more discriminative skeletal representations.

Specifically, we add a shared embedding vector to each corresponding pair of skeleton latent vectors in $\bm{Z}^{src}_p$ and $\bm{Z}^{tgt}_p$, acting as latent-level pairing IDs.
To prevent overfitting to specific bone semantics and promote generalization across skeleton templates, we randomly shuffle the assignment of these embeddings across query points during each training iteration.
Analogous to positional embeddings in vision transformers, this design enhances the distinctiveness of unordered latents, enabling the model to effectively track and transfer local features across poses.

\subsection{Latent Posing Transformer}
\label{sec:transformer}

At the core of our framework is a decoder-only transformer that re-poses shapes within the latent space.
To ensure determinism for supervision, we represent the source shape via channel-wise concatenation of the mean $\operatorname{E}(\bm{Z}^{src}_s)$ and variance $\operatorname{Var}(\bm{Z}^{src}_s)$. 
This representation is concatenated with $\bm{Z}^{src}_p$ and projected into a sequence of pose-aware source input tokens.
Target input tokens are similarly projected from $\bm{Z}^{tgt}_p$.

The source and target tokens are then concatenated into a single sequence and processed by our latent posing transformer.
Through in-context attention, the model implicitly learns the source-to-target shape deformation, completing skinning-free posing in a single forward pass.
Note that this data-driven process is not bound to linear skinning, but can internalize complex deformation behaviors from training data (\eg, preserving volume under twists, detailed in the supplementary material).
Finally, the updated output tokens are projected to predict the mean and variance of the target shape's distribution, from which $\bm{Z}^{tgt}_s$ is sampled for posed mesh reconstruction via SDF decoding and marching cubes.

\subsection{Training with Latent-Space Supervision}
\label{sec:training}

While training regressive models~\cite{jin2025lvsm,yu2025humanram} via end-to-end reconstruction loss is common practice, directly applying SDF supervision to our output shapes yields over-smoothed results lacking geometric detail.
This issue stems from the unordered nature of the latent VecSet. As illustrated in Fig.~\ref{fig:tokens}~(b), the permutation invariance of the output tokens introduces ambiguity into the transformer's input-output correspondence, since any permuted token set decodes to exactly the same shape.
This prevents the model from learning a consistent transformation path, forcing it to average out predictions into a smooth mean result.
To resolve this, we propose a latent-space supervision strategy.

First, we establish a canonical transformation path ensuring that each predicted target latent corresponds to the exact semantic local region of its source counterpart, as depicted in Fig.~\ref{fig:tokens}~(b).
Building upon the one-to-one correspondence (between the target skeleton latents $\bm{Z}^{tgt}_p$ and the source shape latents $\bm{Z}^{src}_s$) established by our dense pose representation (Sec.~\ref{sec:skeleton}), we construct aligned ground-truth target latents $\hat{\bm{Z}}^{tgt}_s$.
Specifically, we encode the ground-truth target shape via the pretrained shape encoder, while setting the query anchor points to $\mathcal{T}(\tilde{\bm{P}})$, \ie, the original source queries $\tilde{\bm{P}}$ transformed by the ground-truth deformation $\mathcal{T}(\cdot)$.
Finally, we supervise the predicted $\bm{Z}^{tgt}_s$ against these aligned ground-truth latents $\hat{\bm{Z}}^{tgt}_s$ using an $L_1$ loss.

\subsection{Fine-tuning with Adaptive Completion}
\label{sec:adaptive_tokens}

The latent-space supervision introduced in Sec.~\ref{sec:training} assumes an isomorphic geometric structure between the source and target shapes, which ensures a one-to-one correspondence in their latent representations.
However, this assumption inherently breaks down during complex character posing, where large deformations reveal previously occluded or fused structures.
For instance, as illustrated in Fig.~\ref{fig:tokens}~(c), lifting an arm exposes the armpit---a surface region lacking latent counterparts in the source mesh due to its initial geometric and topological imperfections.
Representing these newly exposed structures necessitates generating supplementary shape latents on the fly.

Typically, predicting an unordered set of new shape latents relies on indirect geometry-space supervision (\eg, via an SDF loss). This is because establishing direct latent-level correspondence is intractable without knowing the exact spatial query anchors of the required new latents within the VecSet-based representation.
However, geometry-space supervision not only introduces heavy decoding overhead, but also risks disrupting the canonical transformation path established during the first training stage (as discussed in Sec.~\ref{sec:training}).

Instead, we propose an anchor-guided adaptive completion module.
It first predicts a predefined number of 3D spatial anchors for the newly exposed regions based on skeletal and shape context. These anchors are subsequently processed by transformer blocks into adaptive tokens, which are then projected alongside the main target tokens to form a supplementary set of target shape latents.

We formulate the training of these new latents as a bipartite matching problem.
By computing pairwise spatial distances between the predicted and ground-truth anchors, we employ the Hungarian algorithm to establish an optimal one-to-one assignment.
This explicit spatial alignment enables a unified \emph{bipartite-matched latent loss}, which combines an $L_1$ loss to directly supervise the matched shape latents with an MSE loss on anchor coordinates to accelerate spatial localization.

To systematically synthesize training pairs that exhibit such geometric and topological changes, we apply an alpha-wrapping-based remeshing technique~\cite{portaneri2022alpha} to the source shapes while leaving the target shapes clean and intact. This procedure intentionally fuses self-contacting or closely adjacent surfaces, effectively simulating the imperfections (\eg, fused limbs) frequently encountered in AI-generated 3D meshes.
To extract ground-truth data for the newly exposed geometry, we sample points from the un-remeshed target shape to select the specified number of spatial anchors that are farthest from the post-remesh main target query points $\mathcal{T}(\tilde{\bm{P}})$. The corresponding ground-truth shape latents are then extracted using the pretrained shape encoder, similar to the procedure in Sec.~\ref{sec:training}.
Finally, we fine-tune the adaptive completion module while keeping the main latent posing transformer frozen.
This mechanism effectively learns to produce plausible completions, reducing missing-surface artifacts arising from challenging flawed inputs (detailed in the supplementary material).

\section{Experiments}
\label{sec:exp}

\begin{figure*}[t]
    \centering
    \includegraphics[width=1.0\linewidth]{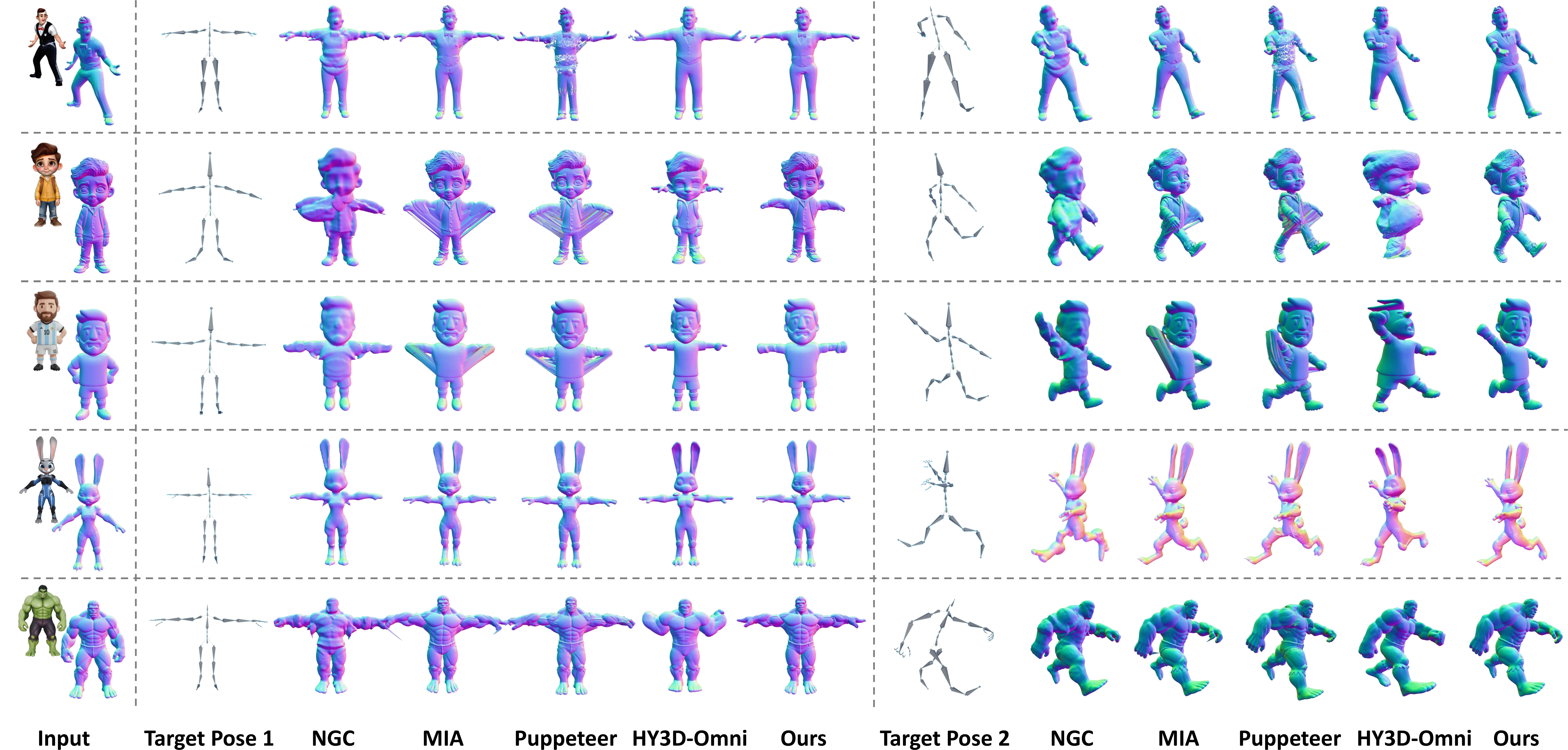}
    \caption{\textbf{Qualitative comparison on diverse characters and poses.}
    We showcase results for re-posing each character into a widely adopted T-pose and an additional random pose.
    Our method produces high-fidelity results across various cases.
    It robustly handles challenging inputs without significant artifacts seen in space-warping~\cite{NGC} or rigging~\cite{MIA,song2025puppeteer} methods, while providing better pose conformance and detail preservation compared to the generative baseline~\cite{hunyuan3domni}.
    }
    \Description{Qualitative comparison on diverse characters and poses.}
    \label{fig:compare}
\end{figure*}

\subsection{Experimental Settings}
\label{sec:exp_set}

\parsection{Data}
Following \citet{MIA}, we prepare our dataset by pairing character models with diverse skeletal poses.
Our character data comprises two main sources: (a) 95 high-quality, artist-designed 3D humanoid models from Mixamo~\cite{mixamo}, and (b) 10k AI-generated humanoid meshes with diverse body proportions from HumanRig~\cite{chu2024humanrig}.
For pose data, we extract skeletal poses from 20k motion sequences from Mixamo~\cite{mixamo}.
Source-target pose pairs are randomly sampled from these collections for training and quantitative evaluation.
As introduced in Sec.~\ref{sec:adaptive_tokens}, we process the source meshes via alpha-wrapping remeshing to simulate common AI-generated topological flaws, while the corresponding clean meshes in the target pose serve as ground truth.
To assess performance on in-the-wild data, we further use \citet{hunyuan3d21} to generate a variety of 3D characters for qualitative comparison.
For quantitative evaluation, we use a full test set of 1k unseen pose pairs and a more challenging 30-pose stress-test subset dominated by proximal limbs and self-contacts, which are both detailed in the supplementary material.

\parsection{Baselines}
We compare our method against representative baselines across different paradigms:
(1) space-warping deformation (NGC~\cite{NGC}),
(2) geometry-space auto-rigging (the latent-space-aligned, re-trained version of MIA~\cite{MIA} for fairness, and Puppeteer~\cite{song2025puppeteer}),
and (3) pose-conditioned generation (HY3D-Omni~\cite{hunyuan3domni}).
Note that Puppeteer is included exclusively for qualitative comparison, as its non-semantic bone structure precludes the precise pose alignment required for valid quantitative evaluation.
While the comparison with HY3D-Omni involves different input modalities, it serves as a valuable paradigm reference to illustrate the inherent limitations of such generative approaches.
Detailed explanations and adaptations required for certain baselines (\eg, providing HY3D-Omni with rendered images) are provided in the supplementary material.

\parsection{Metrics}
We quantitatively evaluate posing quality against the ground-truth shapes using several metrics. Following \citet{zhang20233dshape2vecset}, we compute Chamfer Distance (CD) and F-score on 50k points sampled from the mesh surfaces, while volumetric Intersection-over-Union (IoU) and SDF Root Mean Square Error (SDF-RMSE) are calculated on another 50k points sampled in 3D space.
The average per-sample inference time is also reported.

\subsection{Comparison Results}
\label{sec:compare}

\parsection{Quantitative analysis}
As shown in Table~\ref{tab:comparison}, our method significantly outperforms the baselines in terms of mesh fidelity across all metrics.
Compared to the space-warping deformation~\cite{NGC}, geometry-space rigging~\cite{MIA}, and generative model~\cite{hunyuan3domni}, our approach achieves superior geometric accuracy and surface fidelity.

\parsection{Qualitative analysis}
The qualitative results in Fig.~\ref{fig:compare} further highlight the superiority of our method.
For standard assets, our results are visually comparable to MIA~\cite{MIA}.
However, in challenging cases with non-rest input poses (\eg, limbs in close proximity) or topologically imperfect input meshes, rigging-based methods~\cite{MIA,song2025puppeteer} both produce spiking artifacts due to skinning leakage or fused geometry.
Similarly, the space-warping baseline, NGC~\cite{NGC}, shares this vulnerability to geometry fusing since it explicitly warps the continuous SDF of the defective input topology.
Furthermore, NGC lacks robustness due to its heavy reliance on heuristically defined cylinders (often requiring manual adjustment to achieve the best performance) and generally struggles to preserve high-fidelity surface details.
In contrast, our method remains robust and can adaptively reconstruct the mesh topology to avoid such issues.
The generative model, HY3D-Omni~\cite{hunyuan3domni}, despite producing plausible results, struggles to preserve character identity and details.
Moreover, its sparse pose control often fails on complex poses or out-of-distribution shapes, leading to inaccurate poses, limb misplacement, or even shape collapse.
Thanks to the dense pose representation and the powerful latent transformer architecture, our method consistently produces high-fidelity results that excel in identity preservation, detail accuracy, and pose conformance.

\begin{table}[t]
    \caption{\textbf{Quantitative comparison.} We report the mesh fidelity metrics on the 30-pose stress-test subset characterized by proximal limbs and severe self-contacts.}
    \label{tab:comparison}
    \centering
    \resizebox{1.0\linewidth}{!}{
    \begin{tabular}{l|cccc}
        \toprule
        & CD~$\downarrow$ & F-score~$\uparrow$ &IoU~$\uparrow$ & SDF-RMSE~$\downarrow$ \\
        \midrule
        NGC~\cite{NGC} & 2.75~$\times 10^{-3}$ & 0.7139 & 0.7708 & 0.0484 \\
        MIA~\cite{MIA} & 0.71~$\times 10^{-3}$ & 0.8455 & 0.8602 & 0.0236 \\
        HY3D-Omni~\citepalias{hunyuan3domni} & 4.87~$\times 10^{-3}$ & 0.4238 & 0.4805 & 0.0866 \\
        \textbf{Ours} & \textbf{0.64}~$\times 10^{-3}$ & \textbf{0.8612} & \textbf{0.8724} & \textbf{0.0171} \\
        \bottomrule
    \end{tabular}
    }
\end{table}

\parsection{Identity preservation}
Beyond geometric fidelity, we assess how well the source character's identity is retained after posing, using a perceptual proxy: the DINOv2~\cite{oquab2023dinov2} cosine similarity between rendered views of the source input and the posed output.
Details of this evaluation protocol can be found in the supplementary material.
Apparently, skinning-based methods such as MIA~\cite{MIA} largely preserve identity by deforming the original mesh (their principal failure mode being skinning artifacts), so a more meaningful comparison is against methods that reconstruct geometry.
For the subset in Table~\ref{tab:comparison}, our method attains a score of 0.83 (higher is better), versus 0.78 for HY3D-Omni~\cite{hunyuan3domni} and 0.68 for NGC~\cite{NGC}, quantitatively indicating less identity drift.

\subsection{Generalization Capability}

\begin{figure}[t]
    \centering
    \includegraphics[width=1.0\linewidth]{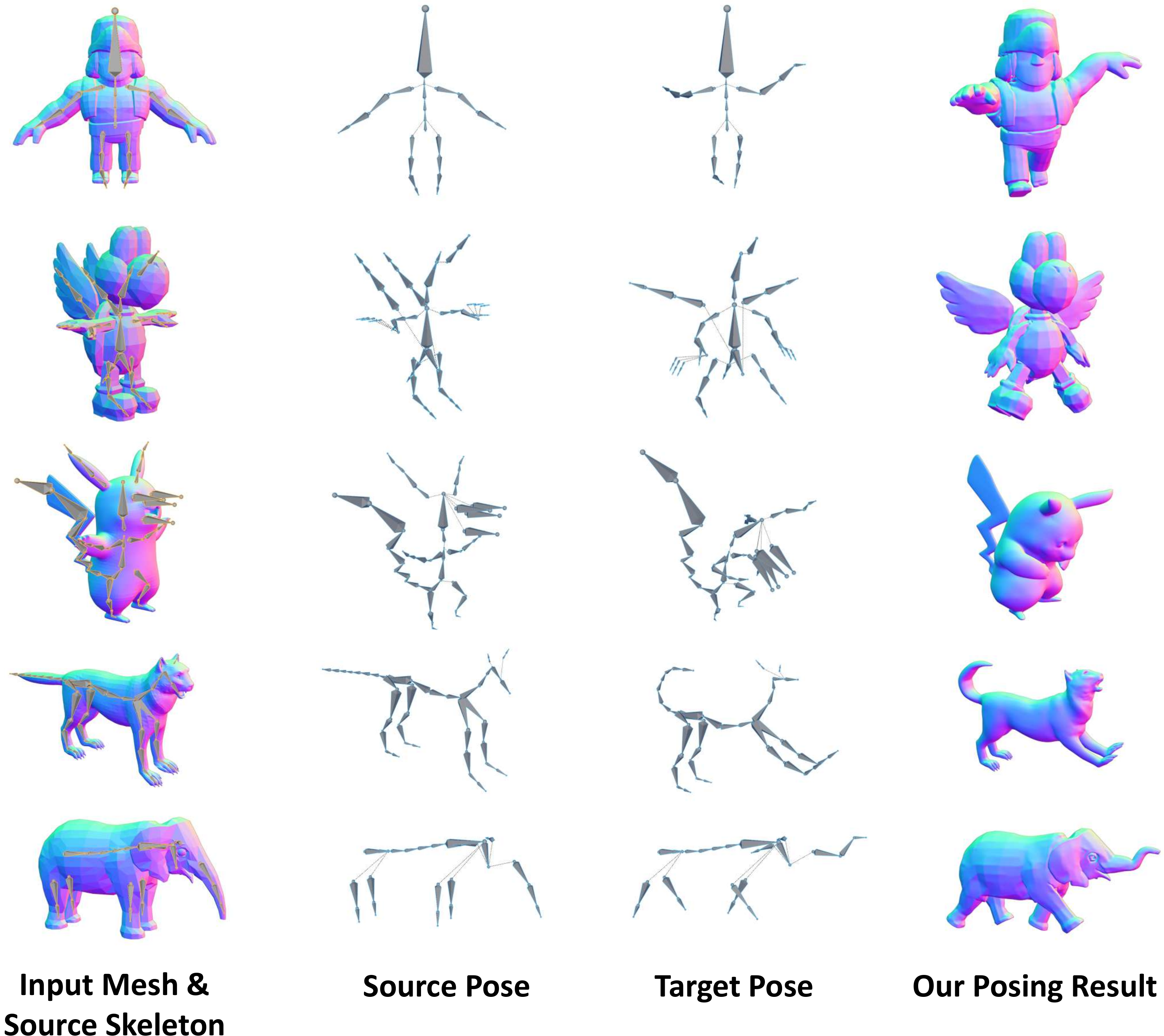}
    \caption{
    \textbf{Zero-shot generalization to out-of-distribution characters.}
    Our method successfully re-poses diverse assets unseen during training, including stylized morphologies, structural accessories (\eg, wings, tails), and quadrupeds.
    }
    \Description{Zero-shot generalization to out-of-distribution characters.}
    \label{fig:ood}
\end{figure}

To evaluate the robustness of our framework, we present zero-shot posing results on out-of-distribution inputs in Fig.~\ref{fig:ood}.
While our model is trained exclusively on standard bipedal humanoids, it also produces plausible results on characters with exaggerated body proportions, novel appendages, and even quadrupedal animals.
This generalizability supports our skeleton-template-agnostic design. It demonstrates that by leveraging a dense pose representation, the network learns generalizable latent-space geometric transformations rather than overfitting to a specific kinematic template.

Moreover, since our training always uses clean skeletons, imperfect input skeletons are also out-of-distribution condition. We analyze this scenario in our supplementary material, which shows that our method tolerates moderate skeletal errors, while severe extraction errors such as extracorporeal bones can still cause failure.

\subsection{Applications}
\label{sec:applications}

\begin{figure}[t]
    \centering
    \includegraphics[width=0.95\linewidth]{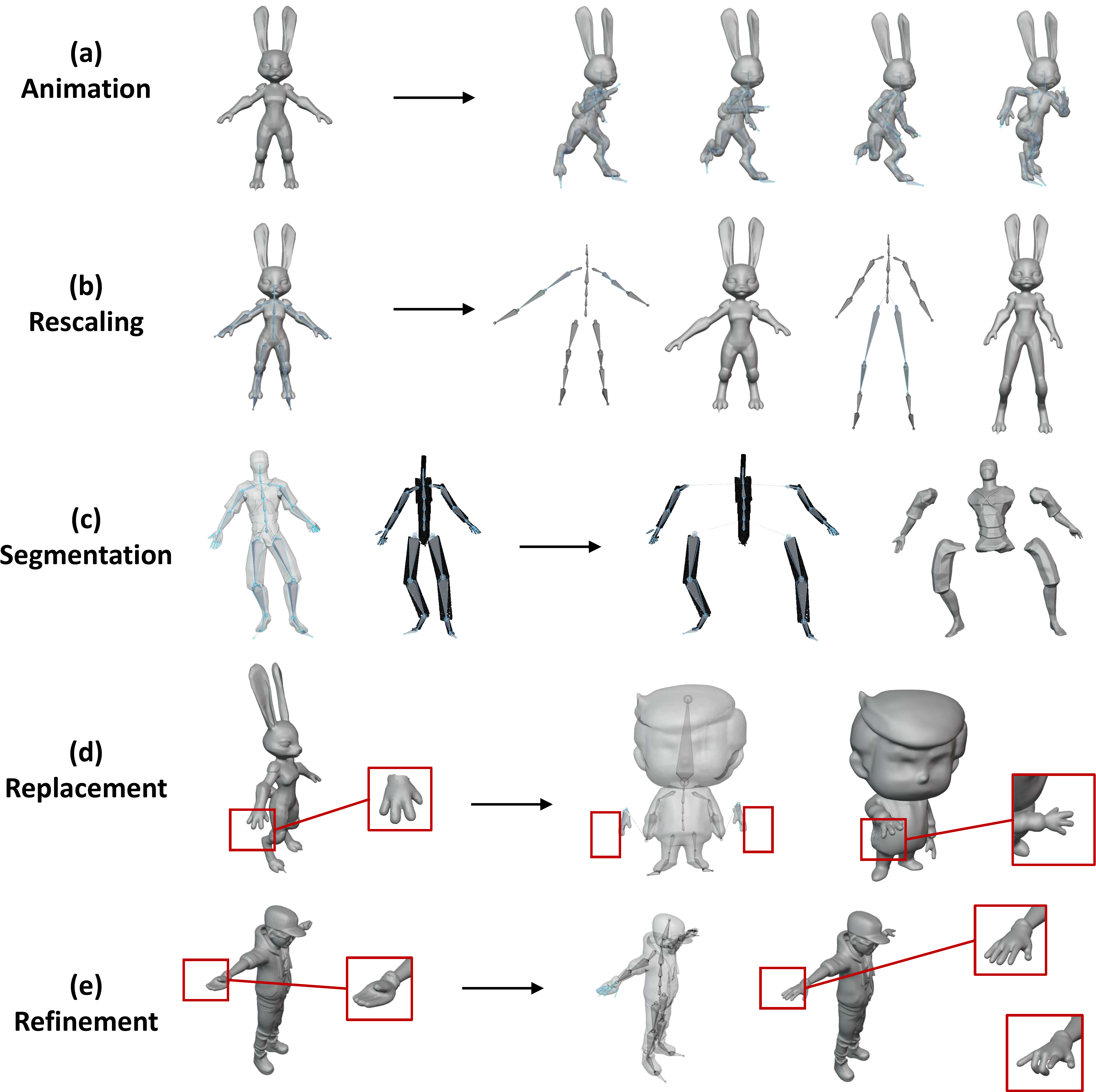}
    \caption{Our method enables various applications, \ie, (a) animation, (b) part rescaling, (c) segmentation, (d) replacement, and (e) refinement.}
    \Description{Our method enables various applications.}
    \label{fig:applications}
\end{figure}

\parsection{Animation}
While our method is designed for static posing, a highly practical application is serving as a robust pre-rigging geometry refiner for downstream animation.
This can be realized by first re-posing a defective input mesh into a clean rest pose with limbs fully extended, and then rigging and animating it using standard pipelines.
This integrates seamlessly with existing graphics workflows while effectively resolving the skinning and topology issues by providing a clean base character.
Notably, despite operating on individual poses independently, our method's robust detail preservation and rapid inference also allow for the direct generation of mesh sequences from target motions, as shown in Fig.~\ref{fig:applications}~(a).
Although not intended for production-level animation, these high-quality sequential outputs provide strong evidence of our consistent high-fidelity posing under continuous observation, serving as a useful tool for rapid prototyping.
We also provide additional sequential results and a detailed consistency analysis in the supplementary material.

\parsection{Part rescaling, segmentation, and replacement}
Our fine-grained and flexible pose representation enables \emph{zero-shot} body part editing, \eg, rescaling, segmentation, and replacement, even without explicit training for these tasks.
These modifications are achieved simply by manipulating the target skeleton.
For instance, character proportions can be effortlessly adjusted by rescaling specific bones, as demonstrated by the limb elongation examples in Fig.~\ref{fig:applications}~(b).
Furthermore, as illustrated in Fig.~\ref{fig:applications}~(c), parts can be cleanly segmented by displacing their corresponding bones far from the main body.
Similarly, foreign parts can be integrated by importing them into the 3D space, removing the original bones, and attaching the new bones to the target skeleton.
Figure~\ref{fig:applications}~(d) shows the character's simplified hands being seamlessly replaced with those of a four-fingered bunny.
Crucially, our latent-space formulation naturally ensures smooth geometric transitions and coherent topology at the boundaries across all these editing operations.

\parsection{Part refinement}
In practice, some character shapes are just abstract geometries lacking detail, such as hands without fingers.
To enable part refinement for them, we fine-tune our model by simply zeroing out the input source shape latents of hands.
As illustrated in Fig.~\ref{fig:applications}~(e), the model then becomes capable of reconstructing cohesive and detailed five-fingered hands guided by the target skeleton. This allows for subsequent fine-grained animation of the fingers.

\subsection{Ablation Study}
\label{sec:ablation}

\begin{table}[t]
    \caption{\textbf{Quantitative ablation results.}
    All variants are evaluated on the full 1k-pose test set.
    We evaluate some variants of our model to validate the effectiveness of our key designs.
    }
    \label{tab:ablation}
    \centering
    \resizebox{1.0\linewidth}{!}{
    \begin{tabular}{l|l|c}
        \toprule
        Module/Stage & Setting & SDF-RMSE~$\downarrow$ \\
        \midrule
        \multirow{2}*{\shortstack{Skeleton\\Encoder}}
        & skeleton as lines (instead of cylinders) & 0.0593 \\
        & w/o discriminative embedding & 0.0345 \\
        \midrule
        Training
        & SDF loss (instead of latent loss) & 0.0442 \\
        \midrule
        \multirow{3}{*}{Fine-tuning}
        & w/o fine-tuning stage & 0.0186 \\
        & latent loss w/o bipartite matching & 0.0240 \\
        & SDF loss (instead of latent loss) & 0.0165 \\
        \midrule
        & \textbf{Ours (full model)} & \textbf{0.0159} \\
        \bottomrule
    \end{tabular}
    }
\end{table}

\begin{figure}[t]
    \centering
    \includegraphics[width=1.0\linewidth]{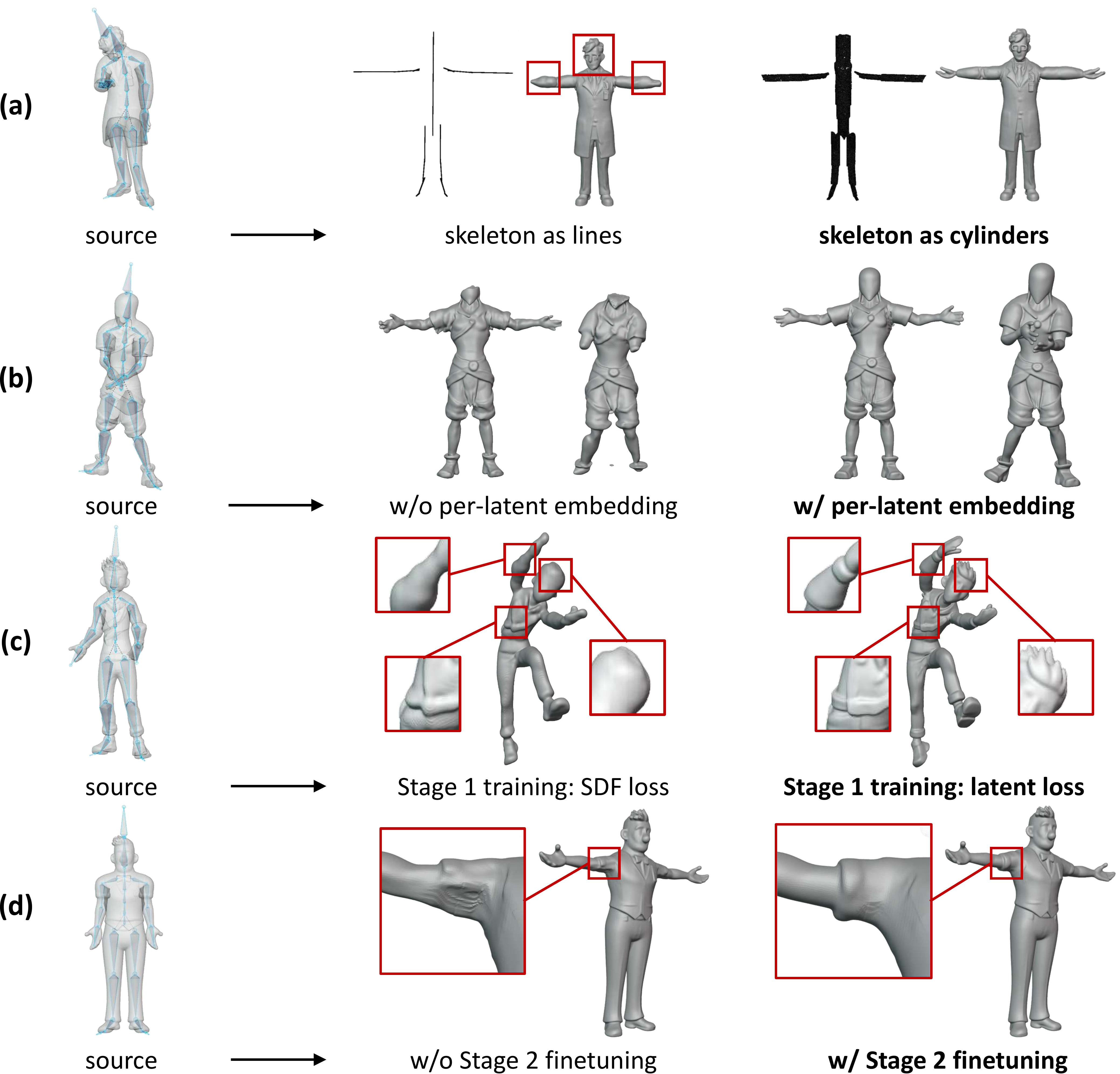}
    \caption{\textbf{Qualitative ablation results.}
    Our designs are crucial for addressing challenges including rotational ambiguity, source-target correspondence, detail preservation, and newly exposed structures.}
    \Description{Qualitative ablation results.}
    \label{fig:ablation}
\end{figure}

We conduct ablation studies to validate our key designs.

\parsection{Skeleton encoder}
We first ablate our pose representation from Sec.~\ref{sec:skeleton}.
As shown in Table~\ref{tab:ablation}, modeling the skeleton as lines significantly degrades performance. This simplification introduces rotational ambiguity along the bone axis, causing artifacts like a twisted head, as seen in Fig.~\ref{fig:ablation}~(a). Our cylinder-based representation resolves this by providing richer 3D context.
We also evaluate the effectiveness of our per-latent discriminative embeddings.
Removing them leads to a substantial performance drop, as the model struggles to establish source-target correspondence, usually resulting in missing limbs (Fig.~\ref{fig:ablation}~(b)).

\parsection{Latent-space supervision}
As discussed in Sec.~\ref{sec:training}, supervising the model in the latent space is crucial.
Training directly with a geometry-space SDF loss yields poor results, producing over-smoothed shapes that lack fine details, as shown in Fig.~\ref{fig:ablation}~(c).
This highlights the importance of our latent loss for preserving high-frequency details.

\parsection{Adaptive completion}
We ablate our adaptive completion module detailed in Sec.~\ref{sec:adaptive_tokens}. Without the fine-tuning stage, the model fails to synthesize newly exposed regions, leading to artifacts like the malformed armpit in Fig.~\ref{fig:ablation}~(d).
Quantitative results further demonstrate that applying a latent loss in this stage is superior to an SDF loss.
Moreover, enforcing a latent loss without our bipartite matching strategy would necessitate defining a strict sequence order for the predicted latents, which is a non-trivial requirement for fundamentally unordered sets. In our ablations, we construct a heuristic baseline that explicitly orders the latents based on the spatial coordinates of their corresponding query anchors. However, this design forces the model to implicitly learn complex, dynamic spatial permutations, rendering the optimization significantly more challenging and yielding results inferior to those achieved with our bipartite-matched latent loss.

\begin{figure}[t]
    \centering
    \includegraphics[width=0.95\linewidth]{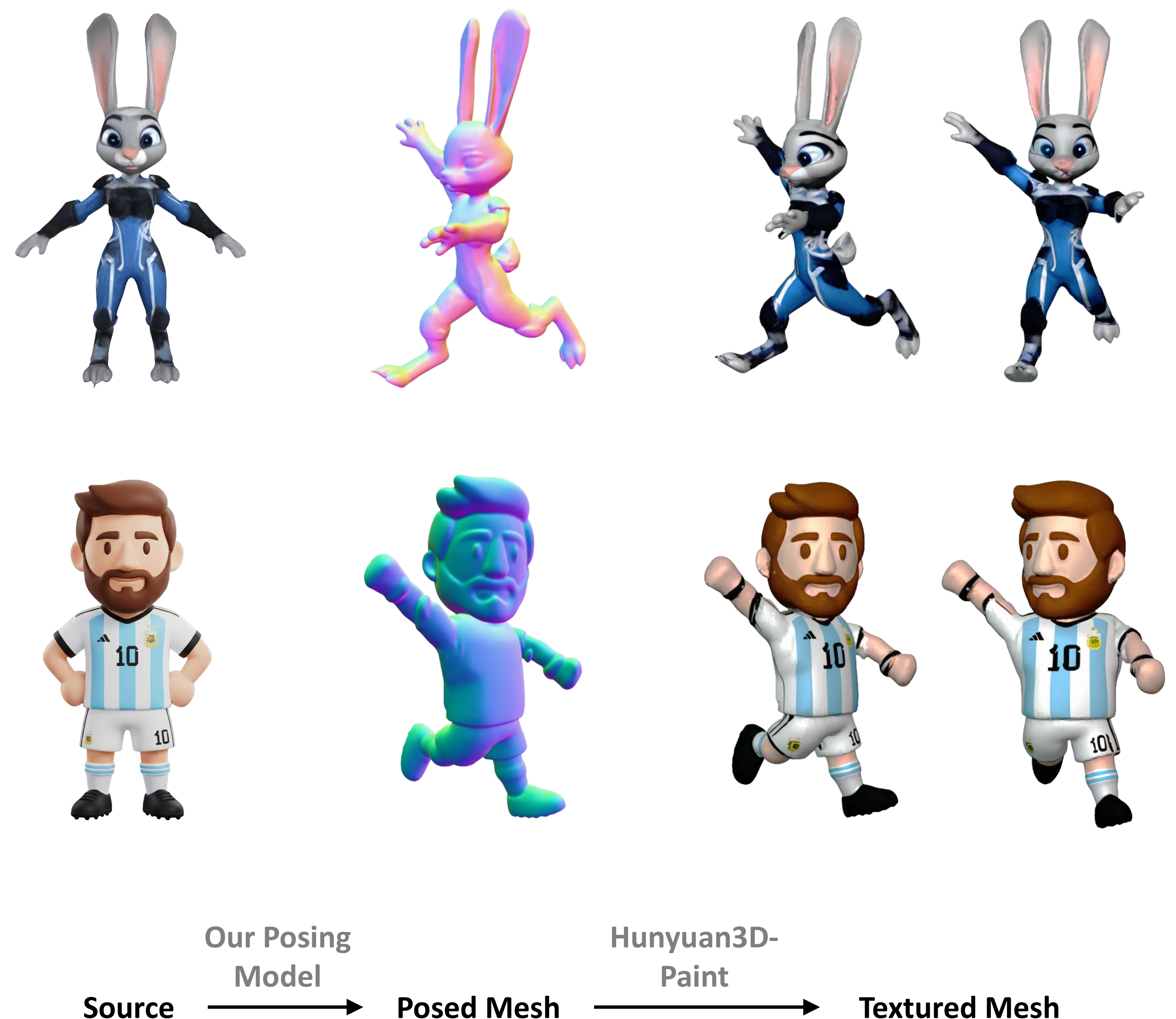}
    \caption{\textbf{Textured results.} We apply textures to our generated posed shapes using Hunyuan3D-Paint~\cite{hunyuan3d21} to demonstrate one feasible solution for realistic rendering.}
    \Description{Textured results.}
    \label{fig:texture}
\end{figure}

\section{Conclusion and Discussion}
\label{sec:conclusion}

In this paper, we present Make-It-Poseable, a novel feed-forward framework that reframes 3D character posing as a latent-space transformation problem. By shifting the deformation paradigm away from the constraints of explicit geometry, our method effectively addresses the limitations of existing techniques, particularly when handling assets with flawed structures.
Comprehensive experiments demonstrate the superiority in posing quality and the promising generalization of our method, alongside its considerable potential for diverse 3D authoring applications.

Despite its merits, our framework still has some limitations that present promising avenues for future work.
First, the geometric fidelity is inherently constrained by the resolution of shape VAE. High-frequency details not captured during the initial encoding process may be lost during latent transformation. While developing new shape representations is orthogonal to our core contributions, our framework can seamlessly integrate future higher-fidelity autoencoders to alleviate this bottleneck.
Second, our framework focuses primarily on geometric manipulation and does not natively preserve or transform appearance. While Fig.~\ref{fig:texture} demonstrates textured results achieved via an external model, integrating material features directly into the latent space to seamlessly pose fully-textured 3D assets remains a promising future extension.
Third, since each posed output is reconstructed via SDF decoding and marching cubes rather than by deforming the original mesh, our method does not preserve the input vertex count, face connectivity, vertex ordering, or UV coordinates, nor does it guarantee temporal consistency across a pose sequence. This is an inherent trade-off of per-pose latent-space reconstruction; accordingly, we position our method as a static pre-rigging geometry refiner rather than a production animation system.

\begin{acks}

This work was supported by the GPU cluster built by MCC Lab of Information Science and Technology Institution, USTC, and the Supercomputing Center of USTC.

\end{acks}

%% file: supplement.tex
\appendix

\section{Implementation Details}
\label{sec:implementation}

\subsection{Shape VAE}

Our 3D shape VAE is adapted from the Hunyuan3D-2.1~\cite{hunyuan3d21} architecture.
The shape encoder processes an input of 32768 points uniformly sampled from the mesh surface, each with its coordinate and normal vector. These points are first downsampled to 4096 via Farthest Point Sampling (FPS). Subsequently, an 8-layer transformer with 16 attention heads performs subset-fullset cross-attention. The resulting 1024-dimensional tokens are then projected into 64-dimensional embeddings for both the mean and variance, producing a VecSet~\cite{zhang20233dshape2vecset} representation of sampled shape latents $\bm{Z}_s \in \mathbb{R}^{L \times d = 4096 \times 64}$.
The shape decoder first projects these latents back to 1024-dimensional tokens, which are then processed by a 16-layer, 16-head transformer using self-attention. Finally, a query-based cross-attention module and an MLP head decode the tokens into Signed Distance Function (SDF) values.
We load the pretrained weights from Hunyuan3D-2.1~\cite{hunyuan3d21} and freeze the entire shape VAE during training.

\subsection{Skeleton Encoder}

As introduced in Sec.~\ref{sec:skeleton}, we represent the underlying skeletal structure as a collection of bone cylinders. To ensure a stable and expressive volumetric proxy, the radius of each cylinder is heuristically set to 25\% of its corresponding bone length.
To generate the skeleton latents, we project the sampled mesh surface points onto these bone cylinders, forming a localized point cloud representation for each bone. This point cloud is subsequently processed by a dedicated encoding network that mirrors the architectural design of the main shape encoder.
Crucially, this symmetrical design enables the direct reuse of the Farthest Point Sampling (FPS) indices extracted during the shape encoding phase. By sharing these spatial anchor indices, we explicitly establish a dense, one-to-one spatial correspondence between the encoded skeleton latents and the shape latents, which strictly aligns the structural and geometric conditions in the latent space.

\subsection{Latent Posing Transformer}

The core latent posing transformer is built upon 24 transformer blocks, featuring a hidden token dimension of 512 and 8 attention heads. To stabilize the training dynamics within deep architectures, QK-normalization is applied across all attention blocks.
The transformer operates on a concatenated sequence of context tokens: the source tokens (projected from the source shape and skeleton latents) and the target tokens (projected from the target skeleton latents).
To efficiently process multiple target poses in parallel, we stack the target token sequences along the batch dimension and duplicate the corresponding source tokens to match this expanded batch size. This parallelization strategy aligns with the implementation practices in \citet{jin2025lvsm}, ensuring highly efficient batched training and inference.

\subsection{Adaptive Completion Module}

The adaptive completion module is designed to infer missing geometric structures through an anchor-guided generation process. The architecture consists of an encoder-decoder transformer structure coupled with an explicit spatial coordinate predictor.

The module takes the concatenated sequence of the encoded shape and skeletal contextual features as input. It first refines these features through an encoder comprising multiple transformer blocks. To determine the spatial locations of the newly exposed geometry, a max-pooling operation aggregates the refined sequence into a single global context vector. This vector is subsequently passed through an MLP to directly regress the 3D spatial coordinates for a predefined number of spatial anchors. In our implementation, we empirically set this number to $512$.

To generate the corresponding adaptive tokens, the predicted 3D anchor coordinates are detached from the gradient graph, processed by a continuous 3D point embedder, and linearly projected to form a set of initial query tokens. These spatial queries are then concatenated with the refined contextual features and fed into a transformer decoder. By in-context attending to the skeletal and geometric cues, the decoder yields the final sequence of adaptive tokens. Finally, these adaptive tokens are passed through an output projector network to map them from the transformer embedding space into the supplementary set of target shape latents.

\subsection{Re-trained Skeleton Extractor}

We adopt MIA~\cite{MIA} as our skeleton extractor, but re-implement it on top of the same frozen Hunyuan3D-2.1~\cite{hunyuan3d21} shape VAE used by our posing model, instead of its original lower-capacity autoencoder based on \citet{zhang20233dshape2vecset}.
This alignment lets the extractor operate directly on the identical shape latents ($L=4096$), avoiding any redundant re-encoding and also improving the accuracy of skeleton extraction, which in turn benefits the subsequent pose-aware latent transformations.
Similar to the original version of \citet{MIA}, we keep the VAE encoder frozen and fine-tune only the decoder's attention layers, along with a set of learnable query tokens and prediction heads that regress blend weights and joint positions. Other training settings are also aligned with \citet{MIA}.

\subsection{Training}

We optimize our framework in two distinct stages.
In the first stage, we train the full model for 5 epochs. The supervision is provided by a direct latent loss ($L_1$ loss) between the predicted target shape latents and the corresponding ground-truth latents, which are deterministically aligned, guided by our dense skeleton correspondence.

In the subsequent fine-tuning stage, we freeze the weights of the main latent posing transformer and encoders/projectors, and specifically train the adaptive completion module for 3 epochs.
Leveraging the optimal one-to-one assignment established via the Hungarian algorithm, we train this module using our unified bipartite-matched latent loss. This objective combines a Mean Squared Error (MSE) penalty on the predicted 3D spatial anchors to facilitate precise localization, and an $L_1$ loss to effectively supervise the supplementary shape latents against their matched ground truth.

Across both training stages, we employ the AdamW optimizer. The learning rate is linearly warmed up to $1 \times 10^{-4}$ over the first 1\% of training iterations, followed by a cosine decay schedule to a minimum of $1 \times 10^{-5}$.
The entire training process takes approximately 120 hours on 4 NVIDIA H200 GPUs.

\subsection{Efficiency}

As a single-pass feed-forward method, our latent posing model offers a clear advantage in efficiency with a per-case inference time of approximately $0.6$ seconds, excluding skeleton extraction (measured on a single NVIDIA H200 GPU with a batch size of 1, including SDF decoding and marching-cubes mesh extraction stages).
Adding the MIA-based skeleton extraction ($\sim$0.45s) gives an end-to-end posing time of about 1 second.
This is comparable to feed-forward auto-rigging~\cite{MIA} and drastically faster than other baselines (HY3D-Omni~\cite{hunyuan3domni} takes $\sim$17s, while NGC~\cite{NGC} requires several minutes for per-case SDF overfitting), imposing minimal overhead on the overall 3D production workflow.

\section{Experiment Details}

\subsection{Data}
\label{sec:suppl_data}

For training, source-target asset pairs are dynamically sampled from 10k humanoid characters paired with 20k Mixamo animation sequences (Sec.~\ref{sec:exp_set}) covering common motions in gaming/3D-modeling workflows, ensuring sufficient pose diversity.
All these assets are used in accordance with their licenses: Mixamo~\cite{mixamo} assets are free to use for research projects under Adobe's terms, and HumanRig~\cite{chu2024humanrig} is released under the Creative Commons Attribution-NonCommercial 4.0 (CC BY-NC 4.0) license.

For quantitative evaluation, our full test set contains 1000 pose pairs, \ie, 10 unseen characters $\times$ 25 unseen motion clips (covering diverse action categories) $\times$ 4 frames per clip at a stride of 30.
While our ablation study (Sec.~\ref{sec:ablation}) uses this full test set, the baseline comparison (Sec.~\ref{sec:compare}) is reported on a more challenging 30-pose stress-test subset, selected before any evaluation by predefined criteria (proximal-limb and severe self-contact cases), which contains the practical failure modes our work targets.

To simulate the topological imperfections common in raw AI-generated assets, such as fused limbs or self-occlusions, we process the source meshes using the alpha-wrapping remeshing technique to generate the input shape, while the original meshes prior to remeshing serve as the ground truth.
Specifically, we use the CGAL-based alpha-wrapping filter~\cite{portaneri2022alpha}, with the alpha and offset parameters set to $0.02$ and $0.001$ of the mesh bounding-box diagonal, respectively. This reliably fuses closely adjacent or self-contacting surfaces, simulating the topological defects.

During training, we augment our data by synthesizing random, anatomically plausible finger poses to compensate for the limited range of finger motion in the original dataset.
Furthermore, following \citet{MIA}, we apply a canonical global transformation strategy to align the orientation of the input shape, which effectively facilitates the pose modeling by simplifying its spatial distribution.

\subsection{Baselines}
\label{sec:suppl_baselines}

Here we provide more details about the baselines adapted and compared in Sec.~\ref{sec:compare}.

\noindent\textbf{Neural Generalized Cylinder}~\cite{NGC} (NGC) is a representative space-warping deformation method that shares the same input/output modality with our method. NGC edits shapes by explicitly warping the spatial query coordinates of a neural signed distance field relative to a set of predefined skeleton-alike generalized cylinders. 
Due to the lack of a generalizable pretrained model, we evaluate NGC by implementing an automatic skeleton-to-cylinder transformation pipeline followed by per-case SDF overfitting using its official codebase.

\noindent\textbf{Make-It-Animatable}~\cite{MIA} (MIA) is a powerful geometry-space auto-rigging method that predicts skeleton and skinning weights for 3D humanoid models of arbitrary shape and pose.

\noindent\textbf{Puppeteer}~\cite{song2025puppeteer} is a recent auto-rigging method for general 3D shapes with an advanced topology-aware skinning predictor. Since it generates non-semantic bones that prevent precise pose alignment for valid quantitative evaluation, we include it only for qualitative comparison, where we manually apply target poses after rigging.

\noindent\textbf{Hunyuan3D-Omni}~\cite{hunyuan3domni} (HY3D-Omni) is a DiT-based 3D generative model capable of producing 3D characters from skeleton conditions. We adapt it for our 3D posing task by providing the rendered image of the textured source shape along with the target skeleton to conform to its required format.
Although this comparison is not perfectly fair due to the different settings, it serves to illustrate the inherent limitations of the pose-conditioned generation paradigm for this task.

\subsection{Identity Preservation Evaluation}
\label{suppl_id}

As discussed in Sec.~\ref{sec:compare}, we use the DINOv2 ViT-B/14 model to evaluate how well the source character's identity is retained after posing.
For each character, we render both the source input and the posed output from three viewpoints (a frontal view and two side views at $\pm 45^\circ$ azimuth), preprocess each image with the standard DINOv2 processor (resize to a shortest edge of 256, center-crop to $224\times224$, and ImageNet normalization), and take the L2-normalized \texttt{[CLS]} token as a global descriptor.
The identity score is the cosine similarity between the source and posed descriptors, averaged over the three views and over all samples in the Table~\ref{tab:comparison} subset.

\subsection{Applications}

Here we provide more implementation details about some of the applications discussed in Sec.~\ref{sec:applications} of the main paper.

\parsection{Part replacement}
Consider the part replacement scenario illustrated in Fig.~\ref{fig:applications} of the main paper, where a source character is to be fitted with new hands imported into the 3D space. Our framework obviates the need for manual mesh editing operations like cutting off the original hands and merging the new ones. Instead, the user only needs to perform two simple steps on the skeleton:
(1) Reposition the original hand bones to align with the new hand geometries, forming a modified source skeleton. It is permissible for these bones to be detached from the character's body (\eg, floating hands).
(2) Roughly align the hand bones with the arm bones to form the target skeleton. This alignment is straightforward as the bones are represented by regular cylinders.
During inference, the association between the hand bones and the new hand geometries ensures that the original hands are automatically ignored and replaced by the new ones in the final posed shape.

\parsection{Part refinement}
The part refinement application is enabled by an additional fine-tuning stage.
During this stage, we override the source shape latents corresponding to the specific body part to be refined (\ie, hands in our experiments) with zeros, while the ground-truth target shape remains unchanged.
As a result, the input tokens for the source part contain only skeletal pose information, with no shape data.
The latent posing transformer learns to interpret this pattern as a signal to recover the missing geometry for the target part, in addition to its primary posing task.
Once fine-tuned, the model can refine the input character's hands by reconstructing well-articulated fingers under the guidance of the provided skeleton.

\section{Additional Experiments and Visualization}

\subsection{Additional Ablation Studies}

\begin{table}[t]
    \caption{\textbf{Additional ablation results.}
    }
    \label{tab:suppl_ablation}
    \centering
    \resizebox{1.0\linewidth}{!}{
    \begin{tabular}{l|l|c}
        \toprule
        Module/Stage & Setting & SDF-RMSE~$\downarrow$ \\
        \midrule
        \multirow{4}*{\shortstack{Skeleton\\Encoder}}
        & displacement embedding & 0.0229 \\
        & per-bone embedding & 0.0317 \\
        & blend-weight embedding (GT weights) & 0.0163 \\
        & blend-weight embedding (predicted weights) & 0.0291 \\
        \midrule
        \multirow{2}*{\shortstack{Latent Posing\\Transformer}}
        & mean-only shape latents (instead of mean \& variance) & 0.0264 \\
        & sampled shape latents (instead of mean \& variance) & 0.0345 \\
        \midrule
        \multirow{4}{*}{Fine-tuning}
        & w/o fine-tuning stage & 0.0186 \\
        & fine-tuning w/o adaptive completion (latent loss) & 0.0182 \\
        & fine-tuning w/o adaptive completion (SDF loss) & 0.0402 \\
        & pre-transformer adaptive tokens (instead of post-) & 0.0168 \\
        \midrule
        & \textbf{Ours (full model)} & \textbf{0.0159} \\
        \bottomrule
    \end{tabular}
    }
\end{table}

We present additional ablation studies to complement those in Sec.~\ref{sec:ablation} of the main paper. The results are summarized in Table~\ref{tab:suppl_ablation}.

\parsection{Skeleton encoding}
We evaluate several alternative embedding variants different from our per-latent discriminative embedding introduced in the skeleton encoder (Sec.~\ref{sec:skeleton}).
In the ``per-bone embedding'' variant, we assign a learnable embedding to each bone, which restricts the framework to a fixed skeleton template.
On this basis, ``blend-weight embedding'' variants mean that the embedding for a given point on the skeleton is computed as a skinning-weighted sum of these per-bone embeddings.
However, these designs are heavily dependent on the quality of the skinning weights. While using ground-truth blend weights to produce embeddings yields comparable scores, this is impractical for in-the-wild inputs.
Other configurations, such as predicted weights or simple displacement vectors, perform worse than our proposed method.
These results confirm that our per-latent embedding offers a robust and effective mechanism for preserving local geometric details across poses.

\parsection{Shape encoding}
Our latent posing transformer manipulates the latent representation of an input 3D shape, which is encoded by a shape VAE. A VAE typically encodes a shape into a distribution characterized by mean and variance vectors, from which the final shape latents are sampled.
Our model utilizes both the mean and variance vectors to represent the shape.
We ablate this design by using only the mean vectors (\ie, zero variance) or the sampled latents. We find that neither variant provides a sufficiently rich representation for effective pose transformation. Both alternatives often result in fragmented mesh surfaces, and in some cases, the marching cubes algorithm even fails to extract a valid mesh from their decoded SDFs.

\parsection{Adaptive completion}
Our fine-tuning approach with adaptive completion effectively addresses the challenge of generating newly exposed structures.
Without the newly introduced adaptive completion shape latents, direct full fine-tuning using a geometry-space SDF loss leads to a collapse in performance, suffering from the transformation path ambiguity issue discussed in Sec.~\ref{sec:training} of the main paper.
Meanwhile, fine-tuning using a latent loss without new latents forces the original shape latents to represent more geometric areas, an ill-posed setting that yields only marginal improvements.
Furthermore, injecting the adaptive tokens after the posing transformer (``post-transformer'') is more effective than injecting them before (``pre-transformer''). This is intuitive because the context required to synthesize the missing structures can be more accurately localized after the pose transformation has been applied.

\begin{figure*}[t]
    \centering
    \begin{minipage}{0.46\linewidth}
        \centering
        \includegraphics[width=1.0\linewidth]{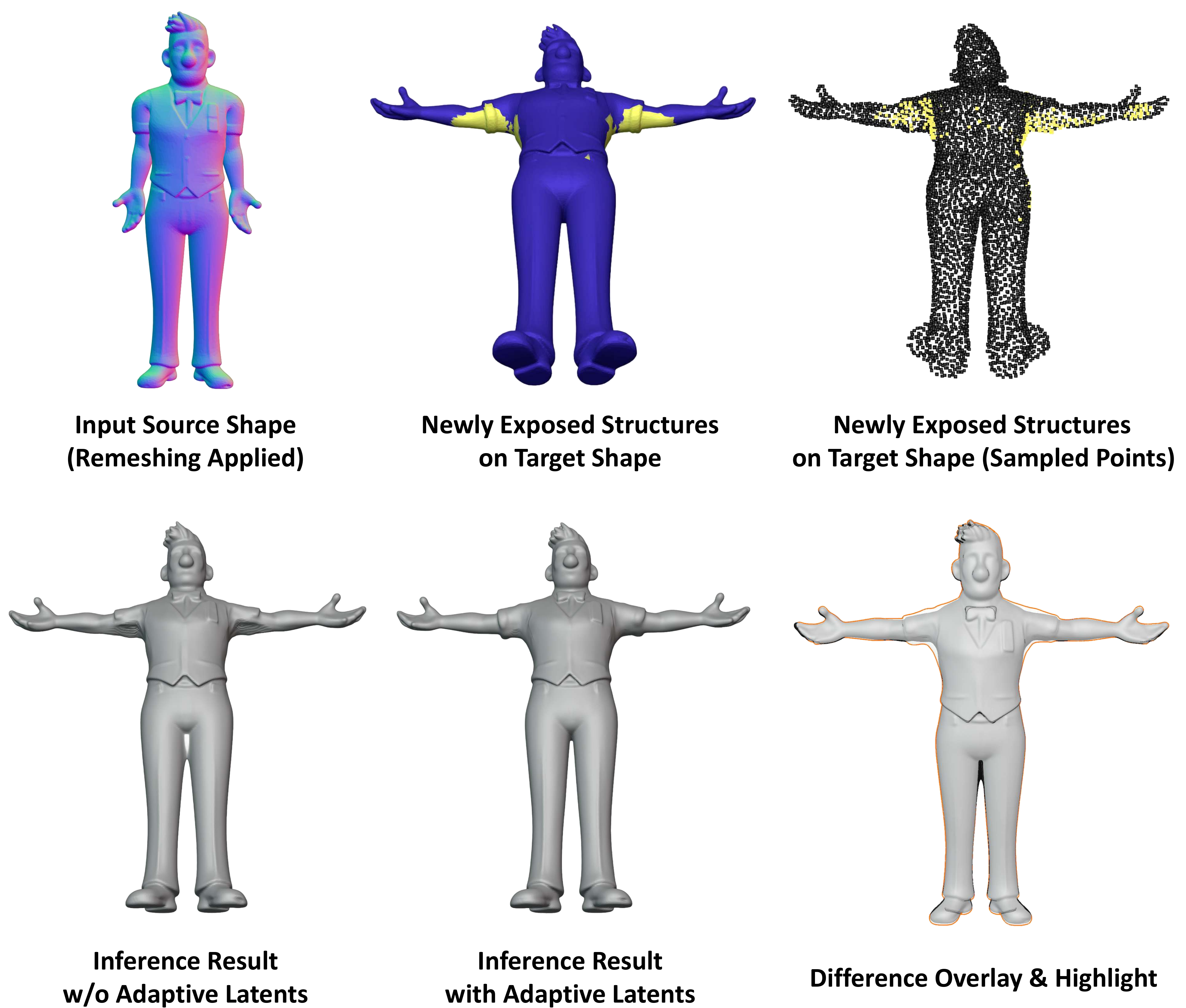}
        \caption{\textbf{Visualization of newly exposed structures.}
        \Description{Visualization of newly exposed structures.}
        \textbf{Top row:} The yellow areas highlight newly exposed surfaces (\ie, armpits) in the target pose that are not sampled from the source shape due to self-contact. Our adaptive completion module is trained to reconstruct these missing regions.
        \textbf{Bottom row:} A comparison of inference results with and without adaptive shape latents. The model with those new latents successfully reconstructs plausible shapes, whereas the model without them fails to produce the complete geometry.
        }
        \label{fig:remesh}
    \end{minipage}%
    \hfill
    \begin{minipage}{0.50\linewidth}
        \centering
        \includegraphics[width=1.0\linewidth]{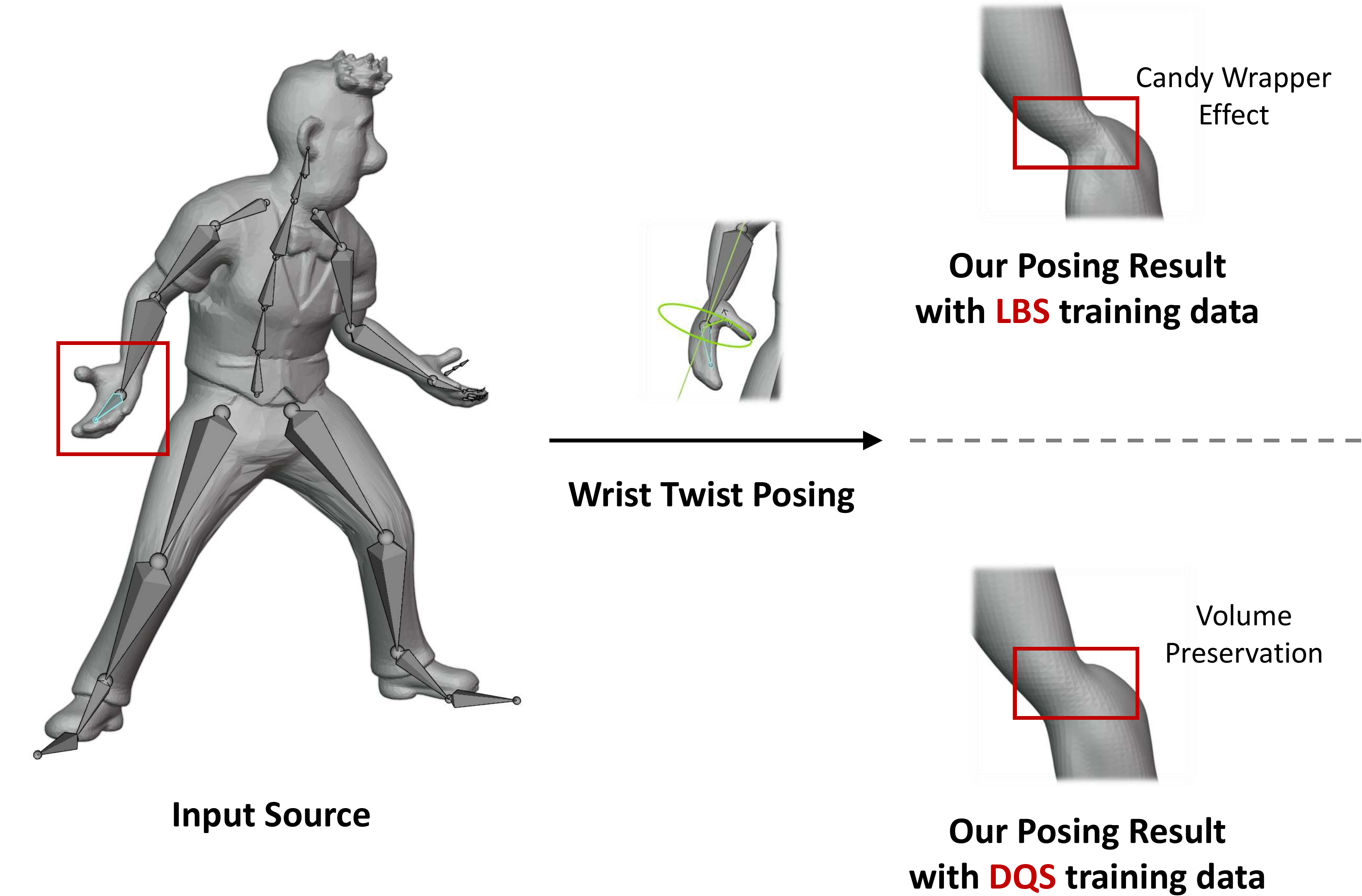}
        \caption{
        \textbf{Data-driven learning of deformations.}
        When posing a severe wrist twist, our model trained on Linear Blend Skinning (LBS) data inherits the well-known ``candy-wrapper'' effect, resulting in noticeable volume loss at the joint.
        Conversely, when fine-tuned on data generated via Dual Quaternion Skinning (DQS), the model successfully learns to preserve the mesh volume. This confirms our framework's ability to internalize and simulate advanced deformation behaviors directly from the training data.
        }
        \Description{Data-driven learning of deformations.}
        \label{fig:dqs}
    \end{minipage}%
\end{figure*}

\subsection{Visualization of Newly Exposed Structures}
\label{sec:remesh}

Figure~\ref{fig:remesh} illustrates the challenge of newly exposed structures that arise when re-posing a character, particularly when limbs are initially in close contact.
Our preprocessing, which involves a remeshing step using the alpha-wrapping algorithm~\cite{portaneri2022alpha}, removes internal or self-contacting surfaces from the source mesh.
Consequently, regions like the armpits are not sampled by the shape encoder, as shown in the top row of Fig.~\ref{fig:remesh}.
This issue is common for in-the-wild 3D character assets, especially AI-generated ones, which often have topological defects like fused body parts.

To address this, we introduce adaptive completion during a fine-tuning stage. The model, already proficient in posing from its initial training, is further guided by a bipartite-matched latent loss during fine-tuning. This effectively directs the adaptive completion module to generate plausible geometry for these previously non-existent surfaces.
The bottom row of Fig.~\ref{fig:remesh} clearly demonstrates the effectiveness of this design.

\subsection{Data-Driven Learning of Complex Deformations}
\label{sec:dqs}

As discussed in the main paper, our feed-forward framework operates in a skinning-free manner during inference. Instead of relying on explicitly defined skinning weights, it implicitly learns the underlying deformation rules directly from the training data. To further validate this data-driven capability, we conduct an additional experiment exploring whether our model can internalize advanced, non-linear skinning behaviors.

Standard Linear Blend Skinning (LBS), which we utilized for our primary training dataset, notoriously suffers from the ``candy-wrapper'' artifact, a severe loss of volume when joints undergo significant twisting motions.
Dual Quaternion Skinning (DQS), on the other hand, naturally preserves volume under such transformations.
To test our model's adaptability, we generate a supplementary training set using DQS and fine-tune a variant of our model on this data.

As illustrated in Fig.~\ref{fig:dqs}, when applying a severe wrist twist, the model trained on LBS data reproduces the volume collapse typical of linear skinning.
In contrast, the model trained on DQS data successfully preserves the wrist volume, perfectly mirroring the DQS deformation characteristics without requiring any modifications to the network architecture.

This experiment explicitly demonstrates that our framework is not inherently bound to linear skinning approximations. By merely altering the training distribution, the network can simulate complex deformation rules.
Consequently, with sufficient high-quality data, our method holds the potential to learn and replicate even more sophisticated deformation behaviors in the future, such as physically-based muscle dynamics or soft-tissue collisions.

\subsection{Results with Texture}
\label{sec:texture}

In a practical geometry-first 3D workflow, texturing naturally follows mesh generation and refinement.
In this work, we primarily focus on shape transformation and leave appearance modeling for future work.
Nevertheless, to demonstrate the potential of our posed shapes for realistic rendering, we provide a feasible solution of applying textures using Hunyuan3D-Paint, a diffusion-based PBR material generation framework (the texturing component of Hunyuan3D-2.1~\cite{hunyuan3d21}).
The results are visualized in Fig.~\ref{fig:texture}.
Hunyuan3D-Paint~\cite{hunyuan3d21} generates albedo, roughness, and metallic maps from multiple viewpoints, conditioned on the 3D shape (via its rendered normal map) and a reference image.
Although this framework is designed for static assets, we observe that the reference image primarily guides the color and style rather than the geometry. Consequently, the character pose in the reference image can differ from that of the input shape while still yielding plausible textured results.

\begin{figure*}[t]
    \centering
    \includegraphics[width=0.9\linewidth]{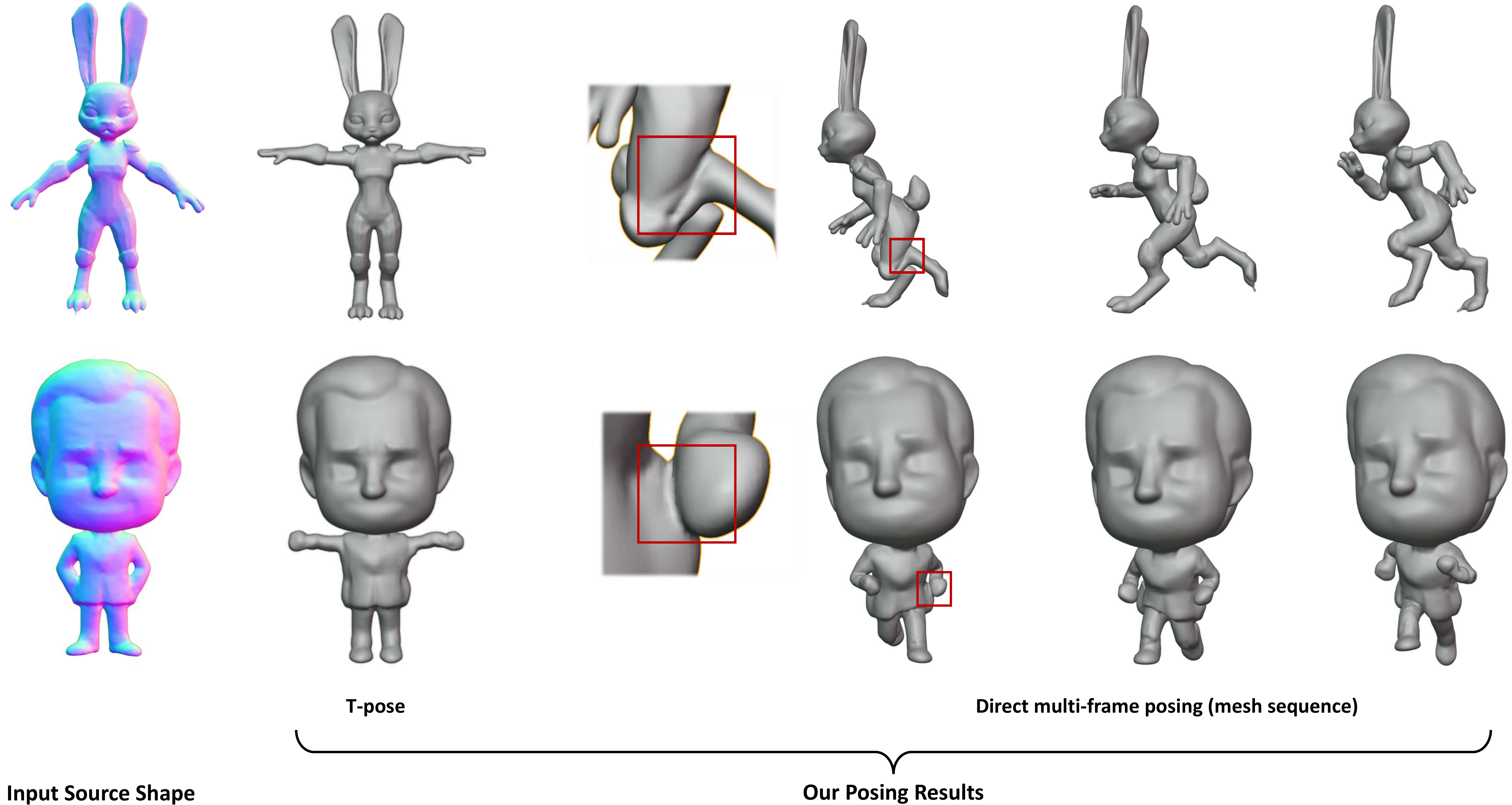}
    \caption{
    \textbf{Direct multi-frame posing and consistency analysis.}
    We pose each input independently into the frames of a motion sequence, without any cross-frame correspondence. The results stay plausible throughout, including the challenging bottom-row character with an exaggerated head-to-body ratio and hands fused into the torso. The close-ups, however, reveal an inherent limitation of our latent posing paradigm, which does not preserve topology, UVs, or temporal consistency across poses. Closely spaced limbs are occasionally merged under certain poses and separated again in others.
    }
    \Description{Direct multi-frame posing and consistency analysis.}
    \label{fig:sequence}
\end{figure*}

\begin{figure*}[t]
    \centering
    \includegraphics[width=0.9\linewidth]{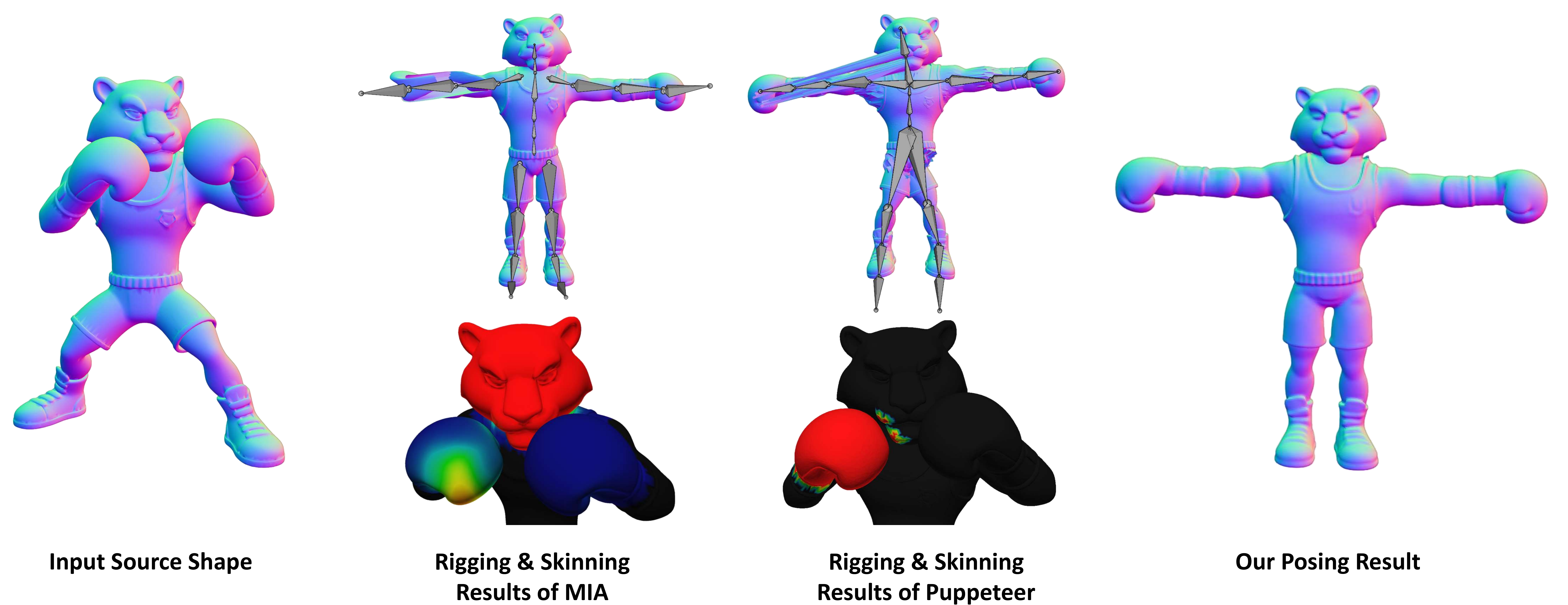}
    \caption{\textbf{Skinning artifacts in geometry-space rigging methods.}
    The right hand of the tiger is positioned very close to its head (without making contact though).
    When re-posed, MIA~\cite{MIA} exhibits skinning leakage, causing some hand vertices to deform unnaturally with the head.
    Puppeteer~\cite{song2025puppeteer} also suffers from noticeable spiking artifacts due to the leaked skinning weights from hand to face.
    In contrast, our method, which is not constrained by skinning, produces a clean posing result.
    }
    \Description{Skinning artifacts in geometry-space rigging methods.}
    \label{fig:skinning}
\end{figure*}

\begin{figure}[t]
    \centering
    \includegraphics[width=1.0\linewidth]{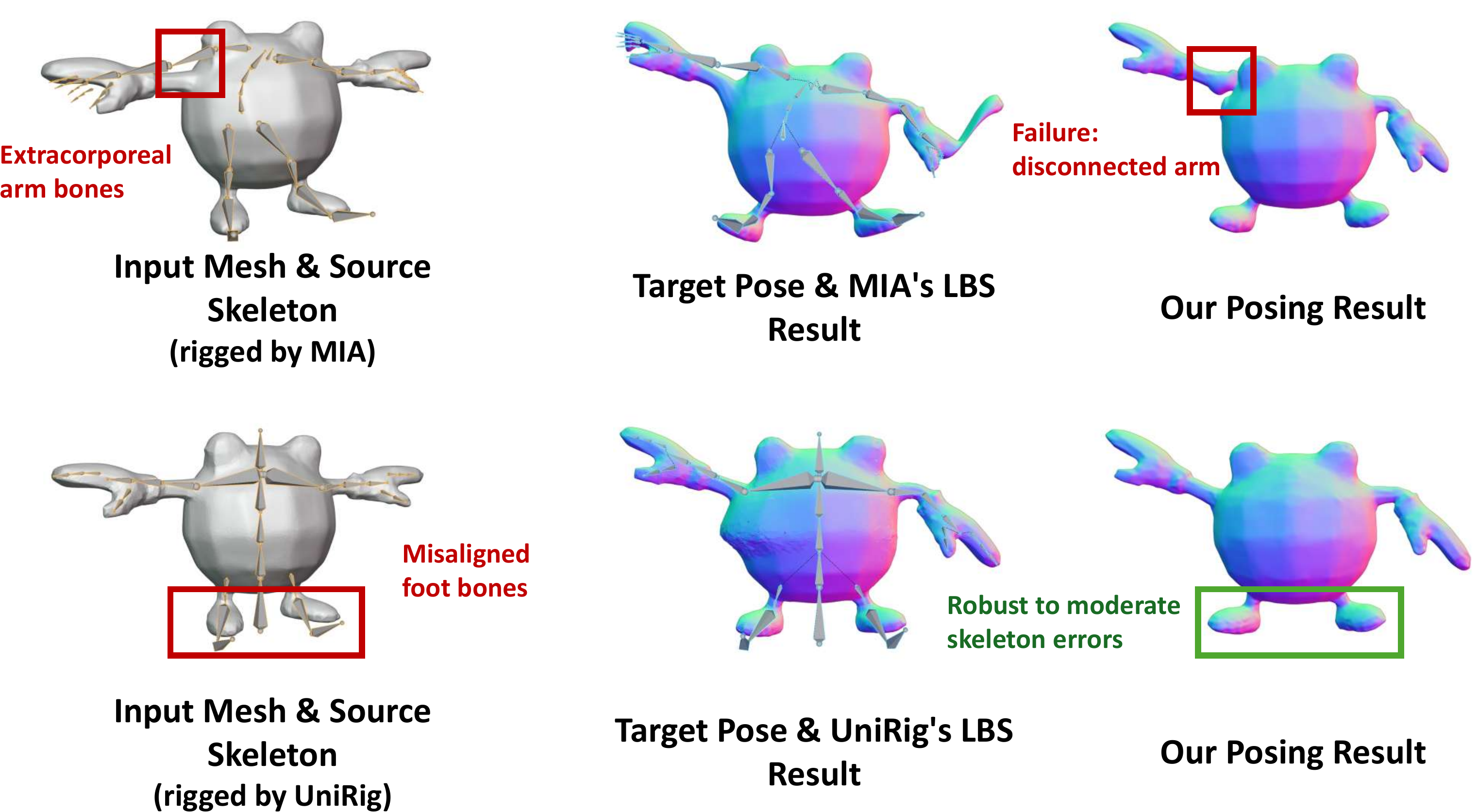}
    \caption{
    \textbf{Failure cases and robustness to rigging prior errors.}
    \textbf{Top:} Severe skeleton extraction errors, such as the extracorporeal arm bones produced by MIA~\cite{MIA}, degrade our dense correspondence and lead to disconnected geometry.
    \textbf{Bottom:} In the presence of moderate prior errors, such as the misaligned foot bones generated by UniRig~\cite{zhang2025unirig}, traditional explicit Linear Blend Skinning (LBS) distorts the mesh. In contrast, our skinning-free posing method remains highly robust, successfully tolerating the input noise to produce plausible articulations.
    }
    \Description{Failure cases and robustness to rigging prior errors.}
    \label{fig:failure}
\end{figure}

\subsection{Sequential Posing and Consistency Analysis}
\label{sec:sequence}

As discussed in the main paper, although our framework is designed for static posing, its robust detail preservation and rapid inference allow it to directly generate mesh sequences from a target motion by posing the input independently for each frame.
Figure~\ref{fig:sequence} presents such results, including a challenging case with an exaggerated head-to-body ratio and hands fused into the torso. The outputs remain plausible across the entire sequence, providing further evidence of our consistent posing quality under continuous observation.

At the same time, this experiment makes the intrinsic limitations of our paradigm explicit. Since each frame is reconstructed independently from its decoded SDF via marching cubes, our method does not preserve mesh topology, UV parameterization, or vertex ordering across poses, and therefore does not guarantee temporal consistency.
As highlighted by the close-ups in Fig.~\ref{fig:sequence}, limbs in close contact tend to be merged during reconstruction under certain poses and separated again in others.
For this reason, we position our method as a static geometry refiner or a pre-rigging tool rather than a replacement for a production animation pipeline. Downstream applications that require a fixed topology across frames can be supported by first re-posing the character into a clean rest pose and then rigging it with standard tools, as described in Sec.~\ref{sec:applications} of the main paper.

\section{Skinning Issue of Geometry-Space Rigging}

Accurate skinning weights are crucial for high-quality character posing in traditional pipelines.
However, even state-of-the-art geometry-space rigging methods struggle to produce satisfactory skinning results, as illustrated in Fig.~\ref{fig:skinning}.

While MIA~\cite{MIA} introduces normal-based geometric features to improve skinning quality and outperforms previous methods~\cite{xu2020rignet, ma2023tarig}, it still suffers from skinning leakage issues that lead to unnatural deformations.
The more recent work, Puppeteer~\cite{song2025puppeteer}, despite employing an advanced skinning predictor that leverages segmentation and mesh-topology priors, still exhibits substantial skinning imperfections like spiking artifacts.
Moreover, while segmentation priors can mitigate skinning leakage in some scenarios, they may also introduce new artifacts due to overly strong assumptions, resulting in rigid deformations and mesh tearing (can be seen in Fig.~\ref{fig:compare}).

Therefore, robustly predicting high-quality skinning weights for arbitrary 3D characters remains an open challenge.
This limitation motivates our exploration of latent-space pose manipulation, which operates in a skinning-free manner.
More crucially, all geometry-based rigging methods are fundamentally unable to correct or handle topological imperfections (\eg, fused limbs or self-occlusions) in the source mesh, a challenge our proposed method robustly overcomes.

\section{Analysis on Robustness to Rigging Quality and Failure Cases}

Since our framework utilizes an automatically extracted skeleton to condition the latent-space transformation, we analyze its sensitivity to the quality of these initial skeletal priors in Fig.~\ref{fig:failure}.

When the predicted skeleton contains severe errors, such as extracorporeal bones entirely detached from the mesh surface (produced by MIA~\cite{MIA}), our dense correspondence mapping becomes unreliable, which leads to geometric artifacts like disconnected limbs.

However, our method exhibits strong robustness against moderate skeletal deviations.
For instance, when provided with misaligned foot bones (produced by UniRig~\cite{zhang2025unirig}), the traditional skinning approach strictly enforces the flawed blend weights, inevitably resulting in noticeable mesh distortion.
In contrast, our skinning-free method allows the network to tolerate input noise and imperfect skeletal guidance, adaptively maintaining shape integrity and yielding plausible posing results even when the prior is suboptimal.

\section{More Discussion and Future Work}

This section further discusses potential extensions of our framework and directions for future research, complementing the limitations outlined in Sec.~\ref{sec:conclusion} of the main paper.

\parsection{Generative completion}
Tasks such as creating newly exposed surfaces or refining absent body parts are not inherently regressive. Currently, our adaptive completion module recovers plausible missing geometry deterministically, relying heavily on statistical patterns observed during training rather than truly generating diverse or novel structures.
Future work could address this limitation by upgrading this component into a generative module. By conditioning the synthesis on multimodal contexts, such as text prompts or reference images, the framework could enable the diverse, user-guided generation of complex structures during posing.

\parsection{Multi-source shape composition}
Currently, the latent posing transformer infers the target shape from a single source mesh.
Extending the framework to accept multiple source shapes, such as assembling individual body parts from different assets or blending features from a gallery of reference poses, could facilitate more flexible and creative 3D authoring workflows through adaptive combination.

\parsection{Broader latent-space editing} 
The principles of our latent posing method could inspire broader applications in general 3D object editing. Beyond skeletal articulation, exploring the use of other dense control signals, such as 3D bounding boxes, could pave the way for more versatile and generalized latent-space shape manipulation.

\parsection{Sparse 3D representations}
Adapting our framework to higher-fidelity spatially grounded sparse representations (\eg, the voxel-based latents in TRELLIS~\cite{xiang2025trellis}) is a promising direction. Such representations may provide stronger explicit spatial anchors than our FPS-based VecSet tokens, but would also require redefining the source-target correspondence when both voxel positions and features change---especially for newly exposed geometry, a case our variable-length VecSet formulation naturally accommodates.